\pdfoutput=1
\documentclass[10pt,twocolumn,letterpaper]{article}
\usepackage[pagenumbers]{cvpr}
\usepackage{multirow}
\usepackage[T1]{fontenc}
\usepackage{microtype}
\usepackage{mathtools}
\newcommand{\method}{\textup{\textsc{Underdraw}}}

\definecolor{cvprblue}{rgb}{0.21,0.49,0.74}
\usepackage[pagebackref,breaklinks,colorlinks,allcolors=cvprblue]{hyperref}

\makeatletter
\patchcmd{\@maketitle}{\vskip .375in}{\vskip 10pt}{}{\PackageError{main}{title patch 1 failed}{}}
\patchcmd{\@maketitle}{\vspace*{24pt}}{\vspace*{16.1pt}}{}{\PackageError{main}{title patch 2 failed}{}}
\makeatother

\title{Underdraw: Curve-Space Completion for Amodal Image Vectorization}

\author{Mengnan Jiang$^{1,2}$ \quad Eva Heckert$^{1}$ \quad Michele Franco Adesso$^{1}$ \quad Antonio Haas$^{1}$ \quad Grace Li Zhang$^{2}$\\[3pt]
$^{1}$Mercedes-Benz \qquad $^{2}$Technische Universit\"at Darmstadt\\
{\tt\small \{mengnan.jiang, eva.heckert, michele\_franco.adesso, antonio.haas\}@mercedes-benz.com}\\[-3pt]
{\tt\small grace.zhang@tu-darmstadt.de}}

\begin{document}
\maketitle
\begin{abstract}
Image vectorization converts raster images into editable vector graphics, yet conventional methods often reproduce only what is visible: shapes cut by a foreground layer stay cut, and the cut shows as soon as that layer is moved. Some recent approaches recover the hidden content by completing it in pixel space and converting the result back to vectors. We instead complete the occluded shapes in curve space. A large-scale census of published SVGs, whose layers retain what designers drew beneath foreground elements, shows that contour truncation and fragment separation dominate real occlusions, that the geometric heuristics we tested recover only a limited share of the hidden shapes, and that a sizable share of occlusion boundaries hide nothing, so forced completion does harm. We therefore build Underdraw around a small language model over SVG commands: conditioned on the visible curves, it predicts the curve segments that span each gap, or returns KEEP to leave the base reconstruction unchanged. To give the model geometric context of the kind it was trained on, a full-image anchor map, predicted in a single forward pass and supervised by designer-placed anchors and the junctions of the planarized visible geometry, provides global anchor cues; a layered fitter combines these cues with visible contours and estimated layer order to recover compact visible paths and the gap endpoints for completion; and constrained assembly integrates the predicted curves into the layered SVG. Generative inference is thus confined to the missing vector geometry, with no pixel-space synthesis and no subsequent raster-to-vector conversion. Experiments show that Underdraw recovers occluded shapes that visible-only vectorization leaves truncated while preserving the fidelity of the visible reconstruction, all within a lightweight raster-to-vector pipeline.
\end{abstract}

\begin{figure}[t]
\centering
\includegraphics[width=\linewidth]{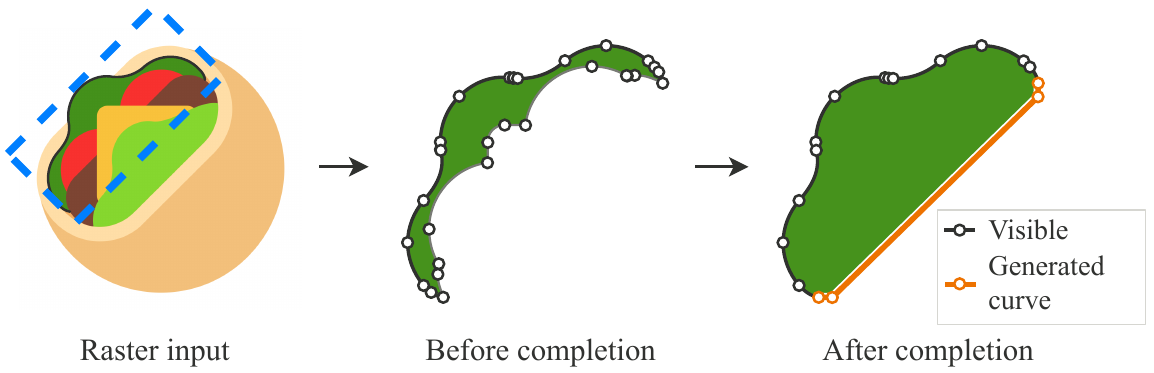}
\caption{\textbf{Complete the shape an edit reveals.} Visible paths provide the context for predicting the missing SVG commands.}
\label{fig:teaser}
\end{figure}

\section{Introduction}
\label{sec:intro}
Image vectorization converts raster images into editable vector paths,
enabling individual shapes to be manipulated independently. Conventional
methods can reproduce the visible composition accurately while leaving
occluded objects truncated. Moving or removing a foreground layer exposes
these incomplete shapes (Fig.~\ref{fig:teaser}). Recovering editable graphics
therefore requires reconstructing both the visible contours and the hidden
geometry that subsequent edits reveal.

Existing approaches offer complementary tools for this problem.
Tracing and optimization recover visible contours accurately
\cite{vtracer,live,adavec}. Geometric continuation connects interrupted
boundaries using local endpoint information~\cite{eulerspiral}. Recent
layered vectorization methods introduce image-generation priors to recover
hidden parts before constructing their vector representation
\cite{layerpeeler,semlayer,amodalsvg}. Meanwhile, large SVG models generate
complete documents from visual or textual input~\cite{starvector,omnisvg}.
Pixel-space completion must re-vectorize synthesized rasters, and
full-document generation can alter observed contours. When much of a shape
is already visible, its reconstructed curves instead provide direct context
for generating only the missing geometry.

We investigate this idea in flat icons and compact emoji. Published SVGs
retain the paths that designers drew beneath foreground elements, making
it possible to study both the structure of real occlusions and the geometry
that resolves them. Across 108,584 files, our census identifies 538,167
significant occlusion events. In 89.46\% of them, a contour is truncated
or split into visible fragments; once the chains are assigned to an object,
both become missing intervals along a closed curve. We address this
missing-interval completion problem. The visible contours do not uniquely determine the hidden geometry. We therefore learn a curve prior to infer the missing intervals from the visible curves and the occluder context.

\textbf{Our key idea is to reconstruct what is seen and generate what is
hidden, in the same curve representation.} We formulate shape completion
as conditional sequence infilling. A small language model reads a curve
sequence with a designated gap and the surrounding occluder geometry, then
predicts the intervening SVG commands, or KEEP to retain the straight
chord between the known endpoints. Updating one interval at a time handles
objects with several gaps.

\method{} turns this operation into a raster-to-SVG workflow: anchor-guided
layered fitting supplies compact visible curves and completion queries, and
constrained assembly integrates the generated intervals beneath the
foreground layers. The completion model operates on B\'ezier paths rather
than front-end features, so it applies to other vectorization front ends
without retraining.

We evaluate hidden-shape recovery, representation compactness, and
appearance after layer editing. On 200 icons with 732 occluded targets,
\method{} achieves 0.414 hidden IoU and 0.192 hidden Boundary IoU,
outperforming LaMa + VTracer (0.397/0.162) and Euler continuation
(0.371/0.170). Its anchor-guided front end uses 76.9\% fewer anchors
than VTracer while achieving lower composite LPIPS. Controlled authored gaps isolate
curve-continuation quality; foreground removal and human ratings assess
the shapes exposed by an edit.
Our contributions are:
\begin{itemize}[leftmargin=*,nosep]
\item A large-scale study of authored SVG occlusions that motivates a
unified missing-contour task and source-referenced evaluation.
\item A 107M-parameter curve language model that generates missing SVG
intervals from visible-curve and occluder context, with a straight-chord
KEEP option.
\item A pluggable completion stage: the same frozen curve model improves
hidden-shape recovery on anchor-guided, VTracer, and AdaVec front ends
without retraining, and
anchor-guided layered fitting with constrained assembly supplies it with
compact context.
\end{itemize}

\section{Related Work}
\label{sec:related}
\paragraph{Reconstructing vector geometry.}
Perceptual fitting balances fidelity, corners, and simplicity
\cite{fidelity,perception,polyfit}. VTracer~\cite{vtracer} traces colored
regions; differentiable rasterization~\cite{diffvg} enables layer-wise and
compact optimization~\cite{live,optreduce,adavec}; learned methods use
unsupervised decoders, superpixels, rounded polygons, or learned anchors
\cite{im2vec,supersvg,vectorark,anchorflow}. These methods reconstruct
visible geometry; we complete the intervals hidden beneath it.

\paragraph{Learning SVG representations.}
Sequence and hierarchical models learn stroke, font, and icon structure
\cite{sketchrnn,svgvae,deepsvg,svgformer}, tokenized generators produce icons
and strokes~\cite{iconshop,strokenuwa}, and multimodal models generate SVGs
from text or images, some with rendering feedback
\cite{starvector,omnisvg,llm4svg,introsvg,duetsvg}. Rather than regenerating a
whole document, our model predicts commands only for a designated gap, with
the supplied curves as fixed context.

\paragraph{Visual priors for vector synthesis.}
CLIP guidance and image-diffusion priors drive vector drawing and sketch
abstraction~\cite{clipdraw,clipasso,clipascene,vectorfusion,diffsketcher,svgdreamer,svgdreamerpp},
alongside implicit, language-guided, and joint vector--raster or latent
models~\cite{neuralsvg,chat2svg,dualdiff,svgfusion}. Unlike these
whole-drawing synthesis methods, we condition generation on observed curves
and complete only occluded intervals, with KEEP when no extension is needed.

\paragraph{Layers and amodal completion.}
Simplification, texture, depth, grouping, and construction stages improve
vector organization and editing
\cite{semantic_simplification,texturevec,easyedit,illustratorsdepth,vectorprism,layertracer}.
Amodal segmentation and image completion infer hidden extent and appearance
\cite{semanticamodal,amodalgroundtruth,lama,pix2gestalt,progressiveamodal,openworldamodal},
beyond local geometric continuation~\cite{eulerspiral}. For vector graphics,
LayerPeeler~\cite{layerpeeler} and AmodalSVG~\cite{amodalsvg} peel and
complete layers as rasters, AmodalSVG restoring natural-image regions with
LaMa and a diffusion inpainter; SemLayer~\cite{semlayer} completes
black-and-white line icons with a fine-tuned pix2gestalt; and
SketchRevive~\cite{sketchrevive} completes pixel sketches. All then
vectorize the pixel result; we instead predict the missing SVG curve commands and
assemble them with the visible paths.

\section{What Occlusion Leaves Missing}
\label{sec:census}
Published SVGs contain both a rendered composite and the underlying
ordered paths. This makes them a record of how designers construct hidden
geometry. We automatically parse their ordered elements and render them
layer by layer, comparing each target's full footprint with the part that
remains visible beneath later layers. Across 108,584 files, this census
identifies 538,167 significant occlusion events; a file can contribute
several events.

\begin{figure}[t]
\centering
\includegraphics[width=\linewidth]{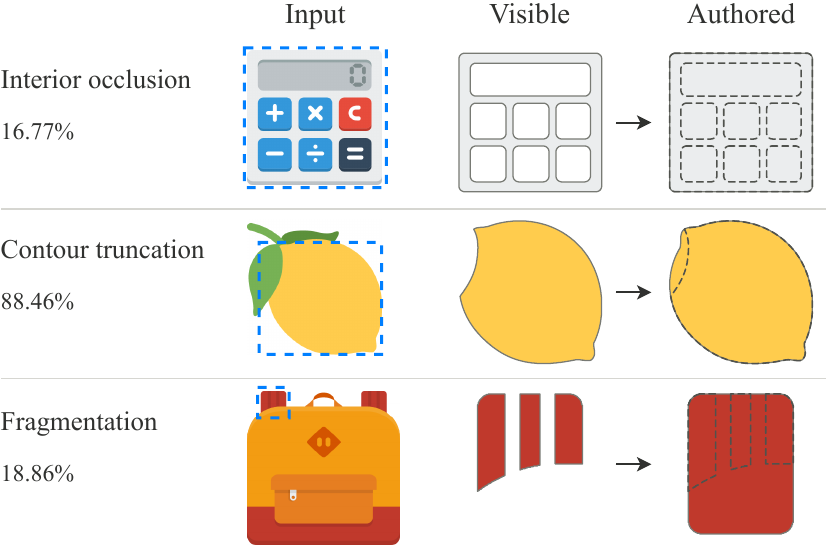}
\caption{\textbf{Occlusion phenomena in authored SVGs.} Dashed outlines mark the originally visible boundary; rates are over 538,167 significant events, and phenomena overlap.}
\label{fig:census}
\end{figure}

\begin{figure*}[t]
\centering
\includegraphics[width=\textwidth]{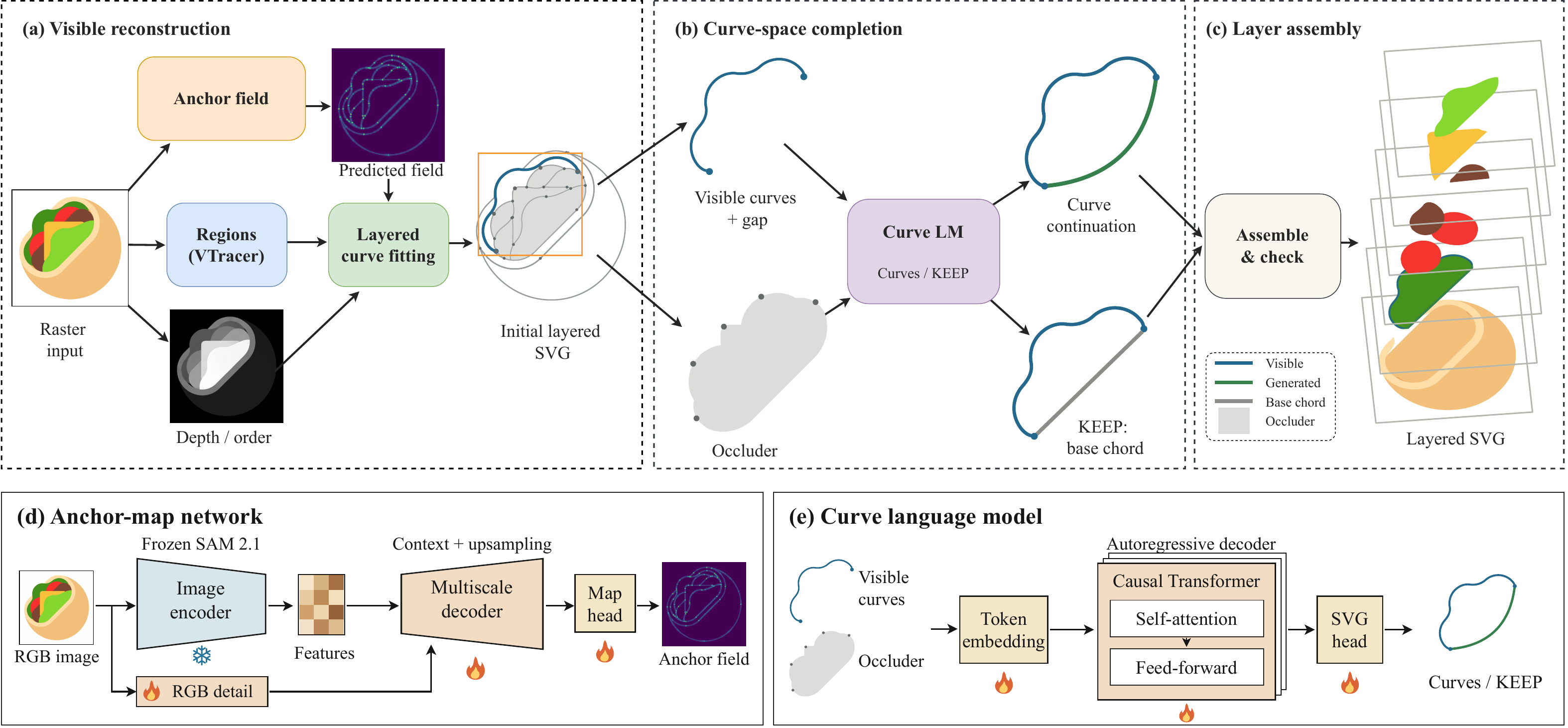}
\caption{\textbf{Reconstruct visible curves; generate hidden intervals.}
\textbf{(a)} An anchor field, raster regions, and depth order guide layered
fitting. \textbf{(b)} Visible curves and occluder geometry condition the
curve LM, which generates missing segments or KEEP. \textbf{(c)} Constrained
assembly integrates them into the layered SVG. \textbf{(d)} Anchor-map
network. \textbf{(e)} Curve model.}
\label{fig:pipeline}
\end{figure*}

\paragraph{Most events interrupt a contour.}
We distinguish interior occlusion, where hidden material is enclosed within
the authored shape; contour truncation, where hidden material touches its
boundary; and fragmentation, where one authored component becomes several
visible pieces. These are overlapping properties, not exclusive classes.
They occur in 16.77\%, 88.46\%, and 18.86\% of significant events,
respectively (Fig.~\ref{fig:census}). Contour truncation or fragmentation
occurs in 481,440 events (89.46\%), counting each event once; most
fragmented targets have at most four visible pieces. Once visible chains
have been assigned to an object, interrupted boundaries admit the same
operation: predict their missing curve intervals. Near- or fully hidden
components (4.45\%) provide little boundary context for curve conditioning.

\paragraph{Filling and contour inference are different problems.}
Interior occlusion can preserve the outer contour. Among 249,356 hole-like
regions with at least 98\% foreground coverage, 95.2\% are filled in the
source. Such regions motivate coverage-aware interior filling, whereas
interrupted outer contours require shape inference. On the 338,895 truncation-only
events, even reference-assisted selection among 17 geometric continuations
reaches hidden IoU $\geq0.8$ in only 20.11\% of cases.

The source path is one authored answer: different hidden curves may
produce the same composite, so we evaluate both agreement with authored
geometry and the result of revealing a completed layer. Many source
boundaries also need no extension: 26.1\% of the 132,626 authored training
queries (Sec.~\ref{sec:lm}) have a hidden interval within 1.5 pixels of
the chord, as do 211 of the 732 whole-image evaluation targets, which motivates the
KEEP option (Sec.~\ref{sec:lm}).

\section{Method}
\label{sec:method}
Given a raster icon $I$, \method{} reconstructs ordered, filled vector paths
whose hidden geometry can be exposed by layer editing. Figure~\ref{fig:pipeline}
shows three stages: recover visible curve context, generate missing
intervals, and assemble the completed paths.

\subsection{From raster regions to curve context}
\label{sec:front}
Let $V_i$ be the observed footprint of region $i$, $\pi(i)$ its estimated
drawing position, and $A_i$ its initial fitted shape. Foreground coverage is
\begin{equation}
O_i=\bigcup_{\pi(j)>\pi(i)}V_j .
\label{eq:occ}
\end{equation}
We use observed foreground regions, so predicted hidden extensions do not
change the coverage used for completion.

\paragraph{Full-image anchor field.}
To reconstruct compact curves with designer-like anchor placement, we
predict a single $1024^2$ field of anchor cues whose peaks guide fitting.
Supervision comes from a planar graph of the visible source geometry:
source paths are clipped by later layers, contours are split at
intersections, and shared boundaries are stored once. Visible source
anchors and junctions define Gaussian peaks, accompanied by a
lower-amplitude boundary band. Local-maximum suppression extracts candidate
anchors; visible contours supply their connectivity. A frozen SAM~2.1
encoder~\cite{sam2}, an RGB detail branch, and a multiscale decoder predict
the field, with 4.4M trainable parameters.

\paragraph{Layered fitting and gap extraction.}
VTracer regions provide initial contours after raster-artifact cleanup.
Illustrator's Depth~\cite{illustratorsdepth} supplies ordering cues,
combined with containment and T-junction relations. A layered fitter
associates the anchor peaks with these contours and fits cubic B\'ezier
paths, constraining visible portions while simplifying hidden boundaries.
The result supplies a base SVG and the context for generation.

Intersecting a fitted boundary with $O_i$ separates visible chains
$C_0,\ldots,C_{k-1}$ and hidden intervals $F_0,\ldots,F_{k-1}$.
Each query connects the end of one visible chain to the start of the next
within the same fitted component. The estimated drawing order determines
the foreground coverage for this interval. The straight chord between
the query endpoints is the \emph{base continuation}: the base reconstruction
closes every gap with it, and KEEP retains it.

\subsection{Conditional completion in SVG command space}
\label{sec:lm}
We initialize a causal Transformer from SmolLM2-135M~\cite{smollm2} and
replace its text embedding and output head with a 533-token SVG
vocabulary, giving approximately 107M parameters. Conditioning on visible
curves and occluder geometry decouples continuation from pixel
appearance.

\begin{figure}[t]
\centering
\includegraphics[width=\linewidth]{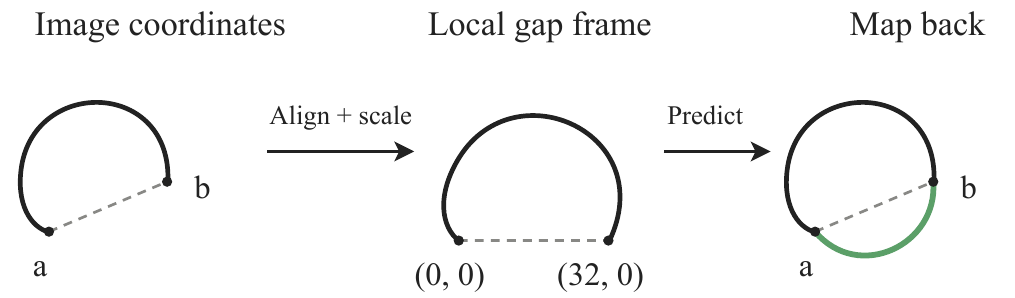}
\caption{\textbf{Predict in a common gap frame.}
The gap endpoints map to $(0,0)$ and $(32,0)$; the predicted interval
(green) is mapped back to the image. Schematic.}
\label{fig:frame}
\end{figure}

\paragraph{A gap-normalized representation.}
Suppose an open chain runs from $b$ to $a$ and the missing interval should
return from $a$ to $b$. We translate the origin to $a$, align the horizontal
axis with $b-a$, and map the endpoint distance to 32 coordinate units.
The perpendicular axis points away from nearby visible chain samples;
ambiguous cases use the occluder-side convention. Quantized coordinates
in this frame (Fig.~\ref{fig:frame}) remove translation, rotation, and uniform scale from the
prediction task. In this frame, a prompt serializes the remaining curve
sequence, a gap marker, and an occluder block that samples the foreground
seam and its geometric depth profile.

The model emits line or cubic B\'ezier commands for the missing interval,
or KEEP. Terminal command variants bind the output endpoint to $b$,
ensuring closure without requiring an exact quantized endpoint prediction.
The known chains retain their floating-point geometry at this stage.

\paragraph{Learning from authored and synthetic occlusions.}
We split source paths at foreground intersections to obtain visible context
and authored hidden intervals. We pretrain on synthetic occlusions, then
fine-tune on synthetic occlusions combined with 132,626 authored queries
from approximately 32,000 sources, 98.1\% from MMSVG-Illustration and
the rest from open Iconify collections. Training minimizes next-token
cross-entropy with extra weight on the first token, which carries the KEEP
decision; targets within 1.5 pixels of the chord on a $1024^2$ canvas are
labeled KEEP.
We use EMA weights and greedy grammar-constrained
decoding. We remove the 24,421 of 164,342 training candidates whose shape
matches an evaluation shape up to similarity or reflection, so no
evaluation image or near-duplicate shape appears in training; see the
supplement.

\subsection{One operation for multiple gaps}
\label{sec:multi}
\begin{figure}[t]
\centering
\includegraphics[width=\linewidth]{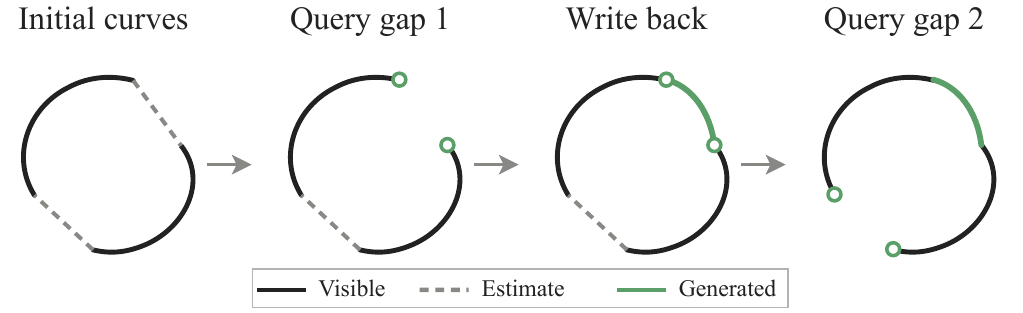}
\caption{\textbf{One missing interval per query.} Each query sees the other gaps' current estimates; a completed gap is written back before the next query. Schematic.}
\label{fig:gap}
\end{figure}

For a fitted component with several hidden intervals, each query removes
one interval and uses the rest of its current path as context:
\begin{equation}
X_j=C_{j+1}\oplus F_{j+1}\oplus\cdots\oplus F_{j-1}\oplus C_j,
\label{eq:context}
\end{equation}
where $\oplus$ concatenates consecutive curves and indices are taken
modulo $k$. Initial hidden intervals
come from the layered fit. We replace $F_j$ with the generated interval
or its base chord on KEEP, then use the updated path for the next query.
We use two cyclic passes; an invalid update retains the previous
estimate (Fig.~\ref{fig:gap}).

\subsection{Constrained assembly into editable layers}
\label{sec:assembly}
Let $P_i$ be the candidate path formed by the known chains and predicted
intervals. We check path validity, sampled self-intersection, and excessive
stray geometry. Accepted proposals contribute only under foreground
coverage:
\begin{equation}
Q_i=A_i\cup(P_i\cap O_i).
\label{eq:assembly}
\end{equation}
Thus $Q_i\setminus O_i=A_i\setminus O_i$ before numerical refitting:
the proposal changes hidden geometry while retaining the initial visible
support. For the candidate $P_i^{\rm ch}$ with every gap replaced by its
chord, the base is $B_i=A_i\cup(P_i^{\rm ch}\cap O_i)$. Both candidates
use the same constrained union, preserving the visible support of $A_i$.

Boolean operations can introduce short segments. We refit these boundaries
with visible-support constraints, preserve paint and drawing order, and
check the final rasterized support. A rejected object falls back, in a
fixed order, to sampled, tangent, and chord continuations, keeping the
first that passes the same checks, or else the initial fit.

\section{Experiments}
\label{sec:exp}
\subsection{Evaluation setting}
\paragraph{Images and authored gaps.}
We evaluate on 200 icons and emoji from Twemoji, FXEmoji, EmojiOne,
Fluent Emoji, and Noto, with 732 eligible hidden targets. We select these
icons manually, discarding those whose authored hidden geometry is clearly
erroneous or ambiguous. Separately, we evaluate 3,062 authored gap queries
extracted from 1,298 SVGs in MMSVG-Illustration~\cite{omnisvg} under a
shared single-gap completion protocol. The first assesses the full workflow, including target
extraction, layer ordering, and gap localization; the second isolates curve
generation.

\paragraph{Metrics.}
\emph{Hidden-shape recovery.} For reference shape $G_i$, foreground coverage $O_i^*$, and
prediction $\widehat G_i$, hidden IoU is
\begin{equation}
\operatorname{hIoU}_i=
\frac{|\widehat G_i\cap G_i\cap O_i^*|}
{|(\widehat G_i\cup G_i)\cap O_i^*|}.
\end{equation}
Hidden Boundary IoU (hBIoU)~\cite{biou} measures contour agreement, and Full IoU
evaluates the complete target. Penalized hidden IoU (p-hIoU) additionally
counts predictions outside both the reference visible region and foreground
coverage as false positives.

\emph{Visible fidelity and compactness.} LPIPS~\cite{lpips} compares the complete
rendered SVG with the input raster; the anchor count, including hidden
geometry, measures representation size.
\emph{Appearance after editing.} After foreground removal, LPIPS, PSNR,
and SSIM~\cite{ssim} measure the revealed region, and CLIP-I~\cite{clip}
compares target crops.

\paragraph{Evaluation protocol.}
Recovery and reveal metrics average over all 732 targets; composite LPIPS
and anchor counts average over the 200 images. Unmatched or failed targets
receive zero geometry scores; reveal metrics are instead computed from the
unedited method rendering for unmatched targets and a white canvas for
missing outputs. Targets are matched to reference layers by
visible-footprint IoU $\geq0.5$. Paired 95\% bootstrap intervals for
the hIoU and hBIoU gains over Euler, tangent, and LaMa + VTracer in
Tables~\ref{tab:story-system} and~\ref{tab:story-controlled} all exclude
zero (supplement).

\begin{table*}[t]
\centering\small
\setlength{\tabcolsep}{2.5pt}
\begin{tabular*}{\textwidth}{@{\extracolsep{\fill}}llrrrrrrrr@{}}
\toprule
& Method & Matched & hIoU$\uparrow$ & hBIoU$\uparrow$
& p-hIoU$\uparrow$ & Full IoU$\uparrow$ & LPIPS$\downarrow$
& \shortstack{Anchors} & Time (s) \\
\midrule
\multirow{4}{*}{\textit{Independent}}
& VTracer & 517 & 0.239 & 0.049 & 0.153 & 0.417 & 0.0091 & 476.2 & \textbf{0.2} \\
& LayerPeeler & 442 & 0.283 & 0.117 & 0.250 & 0.436 & 0.0749 & 339.3 & 109.9 \\
& StarVector-8B & 195 & 0.165 & 0.053 & 0.138 & 0.206 & 0.2344 & 220.6 & 499.6 \\
& OmniSVG-1.1-8B & 145 & 0.108 & 0.033 & 0.088 & 0.146 & 0.2938 & 202.9 & 59.3 \\
\midrule
\multirow{7}{*}{\shortstack[l]{\textit{Shared}\\\textit{front end}}}
& Chord-closed base & 526 & 0.306 & 0.115 & 0.291 & 0.518 & \textbf{0.0059} & \textbf{100.4} & 94.44 \\
& Tangent & 526 & 0.365 & 0.158 & 0.342 & 0.529 & \textbf{0.0059} & 101.9 & 94.44 \\
& Euler spiral & 526 & 0.371 & 0.170 & 0.346 & 0.523 & \textbf{0.0059} & 105.1 & 94.44 \\
& LaMa + VTracer & 526 & 0.397 & 0.162 & 0.364 & 0.528 & 0.0091 & 969.9 & 94.44 + 31.8 \\
& SDXL + VTracer & 526 & 0.183 & 0.080 & 0.171 & 0.468 & 0.0098 & 664.6 & 94.44 + 14.0 \\
& FLUX Fill + VTracer & 526 & 0.305 & 0.120 & 0.277 & 0.464 & 0.0094 & 1604.1 & 94.44 + 134.0 \\
& \textbf{Ours} & 527 & \textbf{0.414} & \textbf{0.192}
& \textbf{0.386} & \textbf{0.546} & \textbf{0.0059} & 110.0 & 94.44 + 2.4 \\
\bottomrule
\end{tabular*}
\caption{\textbf{Whole-image reconstruction} on 200 icons and 732 hidden targets;
unmatched targets score zero. Shared-front-end times list front end +
completion; the geometric rules are not timed separately.}
\label{tab:story-system}
\end{table*}
\subsection{Whole-image layered reconstruction}
\paragraph{Comparisons.}
Table~\ref{tab:story-system} compares visible tracing with VTracer, layered
generation with LayerPeeler~\cite{layerpeeler}, and image-to-SVG generation with StarVector-8B and
OmniSVG-1.1-8B. Shared-front-end baselines replace our completion with
chord closure, tangent continuation, Euler spirals, or LaMa, SDXL
Inpainting~\cite{sdxl}, or FLUX.1 Fill~\cite{fluxfill} followed by
VTracer; chat models are evaluated in Sec.~\ref{sec:large-models}.
SDXL and FLUX.1 Fill run at $1024^2$ with their model-card sampling
settings and one fixed prompt; all configurations and adapters are given
in the supplement.
Because SemLayer~\cite{semlayer} and AmodalSVG~\cite{amodalsvg} have no
public implementations, we evaluate LaMa~\cite{lama} and FLUX.1 Fill, the completion models AmodalSVG builds on.

\begin{figure*}[!t]
\centering
\includegraphics[width=\textwidth]{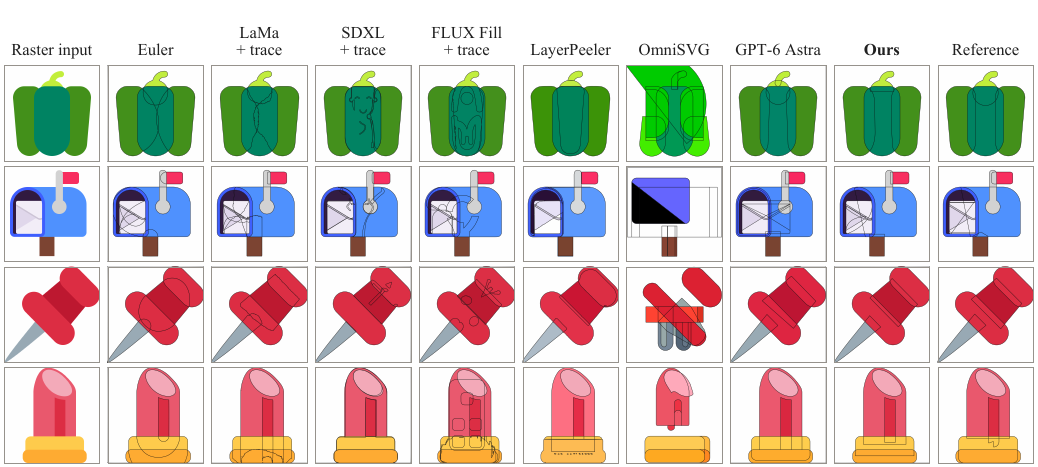}
\caption{\textbf{Whole-image vectorization.}
Solid outlines show all path boundaries, including hidden portions.
SDXL + trace and FLUX Fill + trace replace LaMa with a diffusion inpainter;
GPT-6 Astra generates the layered SVG directly; LayerPeeler as in
Table~\ref{tab:story-system}.}
\label{fig:qualitative}
\end{figure*}

\begin{figure*}[!tp]
\centering
\includegraphics[width=\textwidth]{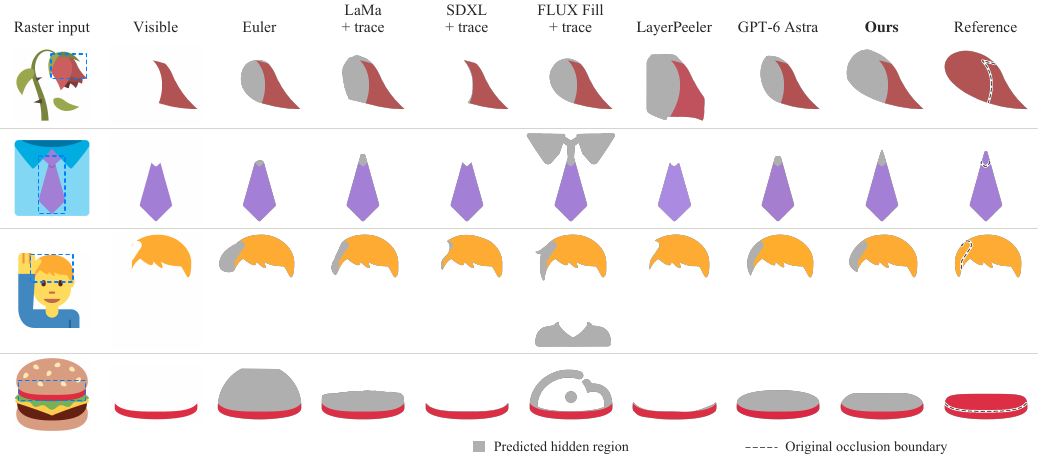}
\caption{\textbf{Revealing completed shapes.}
Objects after foreground removal; gray marks predicted geometry inside the
original occlusion region, dashed lines its boundary on the reference.
Methods as in Fig.~\ref{fig:qualitative}; GPT-6 Astra objects are extracted
from its full-scene SVG.}
\label{fig:approved-examples}
\end{figure*}

\paragraph{Recovering hidden shapes.}
Figure~\ref{fig:qualitative} shows all path boundaries and
Fig.~\ref{fig:approved-examples} the objects after foreground removal.
Euler continuation closes gaps with round arcs and loops (pushpin,
lipstick base), rounds the occluded tip of the tie, and bulges the burger
layer into a dome. Pixel completion followed by tracing produces irregular
geometry: LaMa + VTracer leaves jagged boundaries (mailbox, lipstick base),
SDXL + VTracer stray interior contours (pepper, mailbox), and FLUX Fill +
VTracer spurious shapes (pepper, pushpin, a collar above the tie). OmniSVG
alters the visible composition. \method{} keeps the visible paths and
closes hidden parts with boundaries close to the authored ones, including
the pointed tip of the tie and the flat burger layer. Both geometric rules
derive the whole interval from the endpoints and their tangents, and an
Euler spiral is smooth by construction, so neither can place a corner or
several segments inside a gap, as the curve model's line and cubic
commands can.

\method{} reaches the highest hIoU and hBIoU in
Table~\ref{tab:story-system} (0.414/0.192). Among shared-front-end
baselines, the strongest geometric rule, Euler, reaches 0.371/0.170, and
the strongest pixel completion, LaMa + VTracer, 0.397/0.162;
SDXL and FLUX Fill + VTracer reach 0.183/0.080 and 0.305/0.120.
SDXL often fills the occluded region with background (on controlled-gap
inputs, shape covers 3.6\% of it on average, versus 12.9\% for LaMa), and
FLUX Fill often adds unrelated shapes that also inflate its anchor count.
Among methods with their own front end, LayerPeeler reaches
0.283/0.117, and StarVector and OmniSVG keep fewer input objects
identifiable, matching 195 and 145 targets versus 527 for ours. The
advantage holds on the 525 targets matched by every shared-front-end
method (0.577/0.268, versus 0.554/0.226 for LaMa + VTracer and 0.518/0.237
for Euler) and under p-hIoU, which penalizes excess geometry (0.386 versus
0.364 for LaMa + VTracer).

\begin{figure}[t]
\centering
\includegraphics[width=\linewidth]{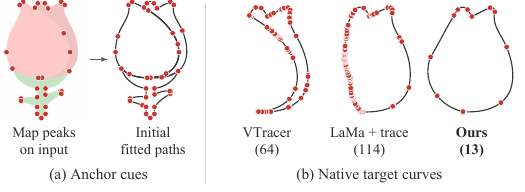}
\caption{\textbf{Global anchor cues and compact native curves.}
(a) Map peaks and initial fitted paths. (b) Target curves with anchor
counts. Red dots mark peaks or anchors.}
\label{fig:anchor-structure}
\end{figure}

\paragraph{Visible fidelity and compact curves.}
In Fig.~\ref{fig:anchor-structure}, the completed target uses 13 anchors,
versus 64 for VTracer and 114 for LaMa + VTracer. Across the 200 images,
\method{} uses 110.0 anchors, 76.9\% fewer than VTracer (476.2), with lower
composite LPIPS (0.0059 versus 0.0091). The anchor-guided
front end keeps visible paths sparse (100.4--105.1 anchors for chord,
tangent, and Euler, also at 0.0059), and curve completion preserves this
budget, whereas pixel completion with VTracer
inflates it to 664.6--1604.1.

\paragraph{Lightweight learned components.}
The learned components are small: a 4.4M-parameter anchor map and the
107M curve model. Completion adds 2.4\,s to the 94.44\,s shared
front end, whereas LaMa, SDXL, and FLUX Fill add 31.8, 14.0, and 134.0\,s
(Table~\ref{tab:story-system});
fitting dominates the front end.

\begin{table}[t]
\centering\small
\setlength{\tabcolsep}{2pt}
\begin{tabular}{@{}lrrrrr@{}}
\toprule
Method & hIoU$\uparrow$ & hBIoU$\uparrow$
& \shortstack{Full IoU$\uparrow$}
& \shortstack{Anchors}
& \shortstack{Time (s)} \\
\midrule
Chord closure & 0.433 & 0.357 & 0.788 & \textbf{13.4} & \textbf{0.6} \\
Tangent & 0.614 & 0.522 & 0.836 & \textbf{13.4} & \textbf{0.6} \\
Euler spiral & 0.646 & 0.546 & 0.830 & 15.0 & 1.1 \\
LaMa + VT & 0.550 & 0.459 & 0.818 & 29.5 & 12.5 \\
SDXL Inp. + VT & 0.473 & 0.400 & 0.793 & 18.2 & 2.5 \\
FLUX Fill + VT & 0.484 & 0.407 & 0.774 & 38.3 & 23.7 \\
\textbf{Ours} & \textbf{0.674} & \textbf{0.585}
& \textbf{0.853} & 14.7 & 2.3 \\
\bottomrule
\end{tabular}
\caption{\textbf{Controlled single-gap completion} on 3,062 queries; time is the
median CPU seconds per query on 100 queries (SDXL and FLUX Fill: mean GPU
seconds per object). VT: VTracer; 59.0\% (LaMa), 70.6\% (SDXL), and 63.0\%
(FLUX Fill) of queries fall back to the chord, versus 5.6\% for ours.}
\label{tab:story-controlled}
\end{table}
\subsection{Controlled single-gap completion}
Every method receives the same visible geometry, endpoints, and
occluders, and only the queried interval is replaced, isolating
continuation from region detection, ordering, and gap localization. \method{} reaches hIoU 0.674 and hBIoU 0.585, compared with
0.646/0.546 for Euler and 0.614/0.522 for tangent continuation
(Table~\ref{tab:story-controlled}). With upstream errors removed, the
learned model thus recovers hidden contours more accurately than geometric
continuation. The margin over Euler is largest
where the geometric families lack expressiveness: on gaps with corners
(0.034 hIoU) and in the top tertile of turning angle (0.044).

Adapted LaMa, SDXL, and FLUX.1 Fill obtain 0.550/0.459,
0.473/0.400, and 0.484/0.407, all between chord closure and tangent
continuation.

\begin{table}[t]
\centering\small
\setlength{\tabcolsep}{2.5pt}
\begin{tabular}{@{}lrrrr@{}}
\toprule
Method & \shortstack{Reveal\\LPIPS$\downarrow$} & \shortstack{Reveal\\PSNR$\uparrow$} & \shortstack{Reveal\\SSIM$\uparrow$} & \shortstack{Crop\\CLIP-I$\uparrow$} \\
\midrule
VTracer & 0.4433 & 16.51 & 0.7801 & 0.8541 \\
LayerPeeler & 0.4062 & 12.65 & 0.7942 & 0.8538 \\
Chord closure & 0.3499 & 19.71 & 0.8202 & 0.8882 \\
Tangent & 0.3362 & 20.30 & 0.8268 & 0.8896 \\
Euler spiral & 0.3397 & 20.37 & 0.8250 & 0.8911 \\
LaMa + VTracer & 0.3366 & 20.25 & 0.8272 & 0.8878 \\
\textbf{Ours} & \textbf{0.3259} & \textbf{20.88} & \textbf{0.8303} & \textbf{0.8956} \\
\bottomrule
\end{tabular}
\caption{\textbf{Appearance after foreground removal} on 732 targets. Reveal
metrics cover the original occluder regions; CLIP-I uses target crops.}
\label{tab:story-editing}

\end{table}
\begin{table}[t]
\centering\small
\setlength{\tabcolsep}{2pt}
\begin{tabular}{@{}lrrrrr@{}}
\toprule
Variant & hIoU$\uparrow$ & hBIoU$\uparrow$
& p-hIoU$\uparrow$ & LPIPS$\downarrow$
& \shortstack{Anchors} \\
\midrule
\multicolumn{6}{l}{\textit{External paths: chord closure / our curve LM}} \\
VTracer + chord & 0.282 & 0.085 & 0.265 & 0.0100 & 311.3 \\
VTracer + \textbf{LM} & 0.359 & 0.138 & 0.330 & 0.0099 & 324.8 \\
AdaVec-40 + chord & 0.261 & 0.065 & 0.241 & 0.0216 & 172.8 \\
AdaVec-40 + \textbf{LM} & 0.316 & 0.100 & 0.289 & 0.0215 & 200.9 \\
AdaVec-10 + chord & 0.252 & 0.063 & 0.234 & 0.0166 & 290.2 \\
AdaVec-10 + \textbf{LM} & 0.316 & 0.107 & 0.289 & 0.0165 & 311.4 \\
\midrule
Chord closure & 0.306 & 0.115 & 0.291 & \textbf{0.0059} & \textbf{100.4} \\
No occluder input & 0.398 & 0.186 & 0.371 & 0.0060 & 111.1 \\
KEEP disabled & 0.413 & 0.189 & 0.385 & \textbf{0.0059} & 110.9 \\
\textbf{Ours} & \textbf{0.414} & \textbf{0.192}
& \textbf{0.386} & \textbf{0.0059} & 110.0 \\
\bottomrule
\end{tabular}
\caption{\textbf{Curve completion across front ends and ablations} on 200 images
and 732 targets with shared assembly. \textbf{LM}: our frozen curve model,
not retrained. External front ends trace per-object visible masks; AdaVec
threshold 10 is native.}
\label{tab:story-ablation}
\end{table}

\begin{figure*}[t]
\centering
\includegraphics[width=\textwidth]{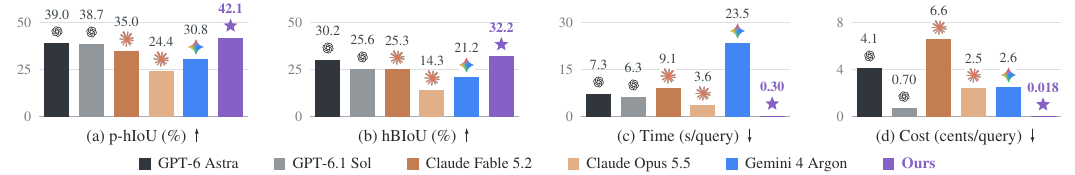}
\caption{\textbf{Curve completion by hosted chat models.} Quality on 20
queries; time and cost on the 15 with complete logs. Our time is measured
on one NVIDIA A100 80GB GPU; our cost uses the on-demand rental price of a
cloud GPU server with that card, \$2.09 per hour.}
\label{fig:chat-paired}
\end{figure*}

\subsection{Quality after layer editing}
\paragraph{Revealed appearance.}
After removing foreground layers, we compare the revealed regions with
the authored references on all 732 targets (Table~\ref{tab:story-editing}).
Although tangent and Euler match our composite LPIPS, \method{} obtains the
best Reveal LPIPS (0.3259 versus 0.3362 for tangent, 0.3366 for LaMa +
VTracer, and 0.4433 for VTracer), PSNR, SSIM, and target-crop CLIP-I.

\begin{figure}[t]
\centering
\includegraphics[width=\linewidth]{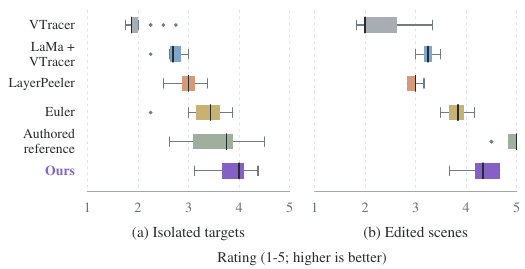}
\caption{\textbf{User ratings} of isolated targets and edited scenes (1--5, higher
is better). Boxes summarize the 18 participants' means; dots are outliers.}
\label{fig:user-ratings}
\end{figure}

\paragraph{Human ratings.}
\label{sec:user-study}
In a blinded study on eight isolated targets and six edited scenes,
eighteen participants rate anonymized results, shown in randomized order
and including the authored reference, for how complete, coherent, and
natural each shape appears (1--5; Fig.~\ref{fig:user-ratings}).
\method{} receives 3.84 for isolated targets
and 4.34 after foreground removal, above all four reconstruction and
completion baselines, the strongest being Euler continuation (3.38/3.82).
The authored references score 3.58/4.91; their low isolated-target score
reflects that designers draw hidden geometry to be covered, not to be
viewed.

\subsection{Completion across curve front ends}
Applied without retraining to anchor-guided, VTracer, and AdaVec paths
on shared visible masks, the frozen curve LM improves over chord closure
on every front end (Table~\ref{tab:story-ablation}), e.g., hIoU from 0.282
to 0.359 on VTracer paths. The improvement holds across anchor
densities: at AdaVec's native threshold, hIoU still rises by 0.064. Anchor-guided reconstruction gives the best final
recovery and visible fidelity while using 110.0 anchors per image,
compared with 200.9--324.8 for the external front ends.

\paragraph{Beyond the fitted base.}
The initial fitted shapes $A_i$ alone, without any gap closure, reach
0.211 hIoU; chord closure (0.306,
Table~\ref{tab:story-ablation}) adds the straight base chord under
foreground coverage through Eq.~(\ref{eq:assembly}), and the complete
system reaches 0.414.
Occluder conditioning adds 0.0155 hIoU.
KEEP matters where no extension is needed: about a quarter of
authored gaps (26.1\% of training queries) stay within 1.5 pixels of the
chord. On 672 such controlled gaps, scored separately from
Table~\ref{tab:story-controlled}, disabling KEEP lowers hIoU from 0.972 to
0.904 and raises the excess area inside the occluder from 5.8\% to 10.7\%
of the hidden area. Tangent and Euler (0.971 and 0.966) also fall below
chord closure (0.999). Because most whole-image targets need completion, the gain averages to
0.003 hBIoU there (Table~\ref{tab:story-ablation}).

\subsection{Comparison with larger generative models}
\label{sec:large-models}

On 20 source-distinct gap queries given visible curves, endpoints, and foreground
summaries, five hosted chat models reach mean p-hIoU 0.244--0.390 and hBIoU
0.143--0.302 (first answer per query; failures score zero), versus 0.421 and 0.322 for our 107M
curve model (Fig.~\ref{fig:chat-paired}a,b). The curve model is also by far
the cheapest: 0.30\,s of GPU inference per query versus 3.6--23.5\,s of
request time, and 0.018 cents per query versus 0.70--6.6 cents of API
charges (Fig.~\ref{fig:chat-paired}c,d). It is thus at least 12$\times$
faster and 39$\times$ cheaper than every hosted model while completing
the curves more accurately.

\section{Conclusion and Limitations}
\label{sec:limits}
\method{} reconstructs visible curves and generates their missing intervals
as native SVG commands through anchor-guided layered fitting, a small
conditional language model, and constrained assembly. No pixels are
synthesized or re-traced at any stage. It recovers coherent
hidden shapes while preserving the visible reconstruction with 76.9\%
fewer anchors than visible tracing, and the frozen curve model transfers
to other vectorizers and completes gaps more accurately, faster, and at
lower cost than hosted chat models. It relies on visible curves and their assignment to
objects, so fully hidden objects, ambiguous fragment grouping, and
incorrect layer order remain challenging. Evaluation covers flat icons and
emoji; extending to textured illustrations is left to future work.

{\small
\bibliographystyle{ieeenat_fullname}
\bibliography{references}}

\clearpage
\maketitlesupplementary
\appendix
\section{Occlusion census details}
\label{sec:census-details}
The census screens 108,584 SVG files, of which 106,604 parse and render
reproducibly; 76,586 of them contain at least one of the 538,167
significant occlusion events. An event is one authored connected
component with at least 12 hidden pixels and a hidden area of at least
2\% of its total area at the census raster resolution. Several events can
come from one SVG. All percentages in this section use events as the
unit unless stated otherwise.

\paragraph{Independent properties.}
Let $K$ be the authored component, $O$ its hidden material, and
$V=K\setminus O$. An event has interior occlusion if a connected part of
$O$ is enclosed by $K$, contour truncation if a connected part of $O$
touches $K$'s boundary, and fragmentation if $V$ has at least two
retained connected components. The three properties are evaluated
independently, so an event can satisfy all three
(Table~\ref{tab:census-joint}). The census uses
8-connectivity, a cross-neighborhood boundary test, and retains visible
components of at least four pixels. Events with fewer than five visible
pixels are counted separately as near- or fully hidden.

\begin{table}[h]
\centering\small
\caption{Joint occurrence counts. The seven rows are disjoint; the
marginal rates in Fig.~\ref{fig:census} overlap.}
\label{tab:census-joint}
\setlength{\tabcolsep}{4pt}
\begin{tabular}{cccrr}
\toprule
Interior & Contour & Fragments & Events & Fraction \\
\midrule
-- & yes & -- & 338,895 & 62.97\% \\
yes & yes & -- & 41,024 & 7.62\% \\
yes & -- & -- & 32,761 & 6.09\% \\
-- & yes & yes & 85,059 & 15.81\% \\
yes & -- & yes & 5,378 & 1.00\% \\
yes & yes & yes & 11,084 & 2.06\% \\
\midrule
\multicolumn{3}{l}{Near / fully hidden} & 23,966 & 4.45\% \\
\midrule
\multicolumn{3}{l}{All significant events} & 538,167 & 100\% \\
\bottomrule
\end{tabular}
\end{table}

\paragraph{Marginal rates and their union.}
Interior occlusion occurs in 90,247 events (16.77\%), contour truncation
in 476,062 (88.46\%), and fragmentation in 101,521 (18.86\%).
Contour truncation and fragmentation overlap in 96,143 events, so
$476{,}062+101{,}521-96{,}143=481{,}440$ events (89.46\%) have at least
one of the two. Among the 101,521 fragmentation events, 59.78\% have two
visible fragments and 87.79\% have at most four. The 338,895
truncation-only events in the first row are the population used for the
geometric-continuation statistic in Sec.~\ref{sec:census}.
The 17 geometric continuations are computed from the visible part and
are placed only under the foreground. Ten are rule-based: an ellipse fit,
a rectangle fit, a horizontal or vertical mirror, the convex hull, a
ring closure, a repeat of a similar visible piece, filling of visible
holes, a bridge between visible pieces, and straight-line and circular-arc
contour continuation. Seven are fitted templates: an ellipse, a rotated
rectangle, a triangle, a five-pointed star, a heart, a mirror about an
axis of any angle, and a $180^\circ$ rotation. Reference-assisted
selection counts an event as recovered when any of the 17 reaches hidden
IoU $\geq0.8$ against the source shape. This holds for 68,154 of the
338,895 events (20.11\%).

At the image level, at least one interior, contour, or fragmentation
event occurs in 49,293, 67,354, and 43,090 files, respectively (46.24\%,
63.18\%, and 40.42\% of the 106,604 reproducible files).
Near- or fully hidden events comprise 22,006 with zero visible pixels and
1,960 with one to four.

\paragraph{Interior filling.}
The 95.2\% statistic concerns 249,356 hole-like regions with at least
98\% foreground coverage. Of these, 237,391 are filled in the source,
2,426 are partly filled, and 9,539 are empty.

\section{Model and inference details}
\label{sec:model-details}

\paragraph{Curve language model.}
The curve model is a causal Transformer initialized from SmolLM2-135M~\cite{smollm2},
whose text embedding and output head are replaced by a 533-token SVG
vocabulary, giving approximately 107M parameters. Its input is a single
prompt in the gap frame below: the remaining curve sequence of the
object, a gap marker, and a 73-token occluder block. Its output is either
the single KEEP token or a sequence of line and cubic B\'ezier commands
whose terminal variant binds the last endpoint to the destination of the
gap. We use the EMA weights and greedy grammar-constrained decoding with
KEEP enabled. The body keeps the SmolLM2-135M configuration: 30
decoder layers, hidden width 576, feed-forward width 1536, 9 attention
heads with 3 key-value heads, rotary position embeddings, and a maximum
context of 8192 tokens; the input embedding and output head are separate
$533\times576$ matrices.

\paragraph{Anchor-map network.}
The anchor-map network predicts one $1024^2$ field of anchor cues from
the RGB input. A frozen SAM~2.1 image encoder~\cite{sam2} supplies features; an RGB
detail branch and a trainable multiscale decoder with a map head produce
the field. The trainable part has 4.4M parameters. Supervision comes
from the planarized visible source geometry: Gaussian peaks at visible
source anchors and junctions, plus a lower-amplitude band along visible
boundaries. At inference, local-maximum suppression extracts candidate
anchors, whose connectivity comes from the visible contours.
The encoder is SAM~2.1 Hiera-B+ applied to the native $1024^2$ input.
Anchor peaks are Gaussians with $\sigma=5$\,px; the boundary band has
amplitude 0.4 and $\sigma=1.5$\,px. The loss weights anchor, band, and
background pixels by 1, 1, and 0.2, with an auxiliary $256^2$ head
weighted 0.4. Training uses 63{,}000 rasterized views (six per icon) of
10{,}500 vector icons and AdamW with weight decay 0.05 and a warm-up
followed by cosine decay. A first stage runs 39{,}375 steps at batch size
24 and peak learning rate $3\times10^{-4}$; its best validation checkpoint
initializes a second stage of 15{,}000 steps at batch size 12 and peak
learning rate $10^{-4}$, in which half of the inputs are downsampled to
0.4--0.8 of their size and upsampled back. Peaks are extracted with
threshold 0.7 and a 7\,px suppression radius.

\paragraph{Endpoint frame and orientation.}
For query start $a$, destination $b$, length $\ell=\|b-a\|$, and
$u=(b-a)/\ell$, choose a perpendicular unit vector $v$. Its sign points
away from the mean of visible-curve samples within distance $\ell$ of
either endpoint; when that signed mean is zero, $v$ points toward the
occluder vertex mean. The coordinate transform is
\[
T(q)=\frac{32}{\ell}\big((q-a)^\top u,\;(q-a)^\top v\big),
\]
so $T(a)=(0,0)$ and $T(b)=(32,0)$. Coordinates are rounded for
tokenization, and decoding applies the inverse transform. This removes
the similarity pose of a nondegenerate gap before quantization. Visible
paths keep their floating-point coordinates.

\paragraph{Occluder conditioning.}
The foreground union uses observed regions above the target in the
predicted drawing order. We keep the exterior ring of each polygon
component; the gap uses the ring containing its chord midpoint, or else
the nearest ring. The 73-token block encodes 16 seam points and eight
depth samples in the same frame as the curve prompt. The depth samples
describe occluder geometry, not depth-network predictions. If the
foreground union cannot be converted to polygons, the occluder of the
individual query is used instead.

\begin{figure*}[t]
\centering
\includegraphics[width=\textwidth]{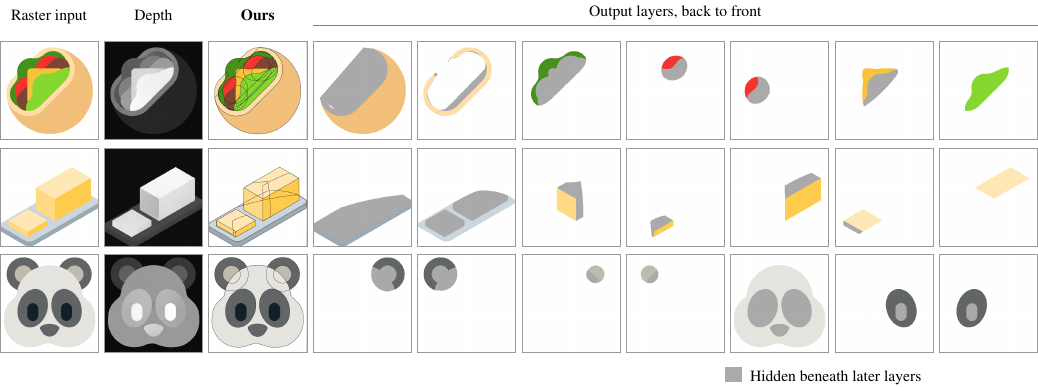}
\caption{\textbf{Depth cue and separated output layers.} Brighter depth
is closer to the viewer. \method{} is shown with all path boundaries; the
seven largest output layers follow in drawing order, with the part
hidden beneath later layers in gray.}
\label{fig:layers-depth}
\end{figure*}

\paragraph{Layer order and output layers.}
Illustrator's Depth~\cite{illustratorsdepth} predicts a depth map from the raster, which is
combined with containment and T-junction relations to order the regions.
Figure~\ref{fig:layers-depth} shows the depth maps next to the output
layers separated in drawing order: each layer is a closed path that
continues beneath the layers drawn after it.

\paragraph{Gap updates, assembly, and fallback.}
Other gaps initially contribute their fitted hidden segments as context.
For a candidate $P$ and foreground mask $O$, assembly unions the initial
fitted shape $A$ with $P\cap O$ and then refits hidden boundaries; the
chord-closed base uses the same operator with straight gap chords. Each
KEEP decision restores its chord, including in the second cyclic pass;
an inference error retains the previous gap estimate.

Candidates must be valid, non-self-intersecting paths (checked on
samples), and their area outside $A\cup O$ may not exceed the area of
$A$. Hidden refitting protects nearby visible support and model-derived
segments within 1.5 pixels. A final raster check allows at most
64~pixels$^2$ of lost visible support on the $1024^2$ canvas. A rejected
object tries, in order, eight samples at temperature 0.7 with KEEP
disabled, then tangent completion, then chord closure, each under the
same checks; the first passing candidate is accepted, with no
reference-based ranking. If none passes, or no completion query can be
formed, the initial fitted object is kept.

\section{Training data and optimization}
\label{sec:training}
\paragraph{Authored data and evaluation filtering.}
The authored pool contains 132,626 queries from approximately 32,000
sources: 130,112 (98.1\%) from MMSVG-Illustration and 2,514 from open
Iconify collections. We retain closed contours with simple opaque fills.
The pool is filtered from 164,342 candidates. Removing candidates whose
shape matches an evaluation shape removes 24,421; identical prompts
remove 91, duplicate rows 1,467, and capping each near-shape class at
eight rows 5,737, for 31,716 removals in total. The shape comparison is
invariant to similarity transforms, reflection, traversal direction, and
start-point shift. Twemoji, EmojiOne, and FXEmoji are additionally excluded as whole
libraries. Authored data added during fine-tuning passes the same
geometric filter, together with library and file-name exclusions.

\paragraph{Mixture.}
Fine-tuning uses 3,038,400 sample presentations over 6,830 updates
(Table~\ref{tab:training-mixture}). Presentations count repeated and
augmented examples, not unique shapes. Real-shape overlays place real
occluders over real target shapes to create additional large gaps;
synthetic examples include single-gap and two-fragment cases. ``Other''
collects small additional authored and synthetic groups.

\begin{table}[h]
\centering\small
\begin{tabular}{lrr}
\toprule
Source group & Presentations & Fraction \\
\midrule
Authored pool & 651,998 & 21.5\% \\
Synthetic single-gap & 997,025 & 32.8\% \\
Synthetic two-fragment & 911,520 & 30.0\% \\
Regular synthetic shapes & 177,004 & 5.8\% \\
Real-shape overlays & 223,500 & 7.4\% \\
Other & 77,353 & 2.5\% \\
\midrule
Total & 3,038,400 & 100.0\% \\
\bottomrule
\end{tabular}
\caption{Fine-tuning sample presentations (rounded fractions).}
\label{tab:training-mixture}
\end{table}

\paragraph{Augmentation and labels.}
Augmentation includes traversal reversal, shape deformation, anchor
split, merge, refit and slide operations, and occluder erosion or
dilation; the later repetitions of authored examples use stronger shape
and anchor augmentation. Similarity transforms cancel in the gap frame
and are not used. KEEP labels are recomputed after augmentation: a target
is KEEP if its hidden interval stays within 1.5 pixels of the chord on
the $1024^2$ canvas. The occluder block is dropped for 15\% of single-gap
examples, so the model also works without it. Each authored query is
used at most six times.

\paragraph{Optimization.}
Synthetic pretraining runs for 203,000 updates. Fine-tuning uses AdamW,
gradient clipping at 1.0, and EMA decay 0.999. A batch has 320 examples
(224 single-gap, 96 two-fragment) for the first 1,500 updates and 480
(336 and 144) afterwards, so
$1{,}500\times320+5{,}330\times480=3{,}038{,}400$. The first-token loss
weight is 3 for the first 2,000 updates and 5 thereafter; other target
tokens have weight 1. The peak learning rate is $5.657\times10^{-4}$,
with warmup, a constant stage, and linear decay to zero over the final
330 updates. We evaluate the final EMA weights.

\section{Evaluation details}
\label{sec:eval-details}
\paragraph{Collection and matching.}
The whole-image evaluation uses 200 manually selected icons and emoji
(75 from Twemoji, 61 from FXEmoji, 42 from EmojiOne, 20 from Fluent
Emoji, and two from Noto) with 732 reference targets. Each target
is matched to an output path by visible-footprint IoU $\geq0.5$.
Recovery metrics weight targets equally, and unmatched or failed
targets score zero on every geometry metric. Pixel-completion baselines
replace only the objects that form completion queries; other objects
keep the shared initial fit. Intervals are paired, image-clustered
bootstrap intervals with 10,000 draws and seed 0.

\paragraph{Mask metrics.}
Let $T$ be the reference target, $O$ the reference foreground union,
$V=T\setminus O$, $H=T\cap O$, and $P$ the matched prediction.
Penalized hidden IoU is $\text{p-hIoU}=\operatorname{IoU}(P\setminus V,H)$;
unlike hIoU, it counts predicted area outside both $V$ and the occluded
region as false positives. Boundary IoU uses an interior
boundary band whose width is 2\% of the image diagonal (28.96 pixels);
hidden Boundary IoU removes $V$ from the prediction and target bands
before computing their IoU.

\paragraph{Paired intervals for Table~\ref{tab:story-system}.}
Table~\ref{tab:paired-wholeimage} gives the paired gains of \method{}
over each method in Table~\ref{tab:story-system} that shares our front end,
and over VTracer and LayerPeeler. All intervals exclude zero, including
those over Euler, tangent, and LaMa + VTracer.

\begin{table}[h]
\centering\footnotesize
\setlength{\tabcolsep}{1.5pt}
\begin{tabular}{@{}lrrr@{}}
\toprule
Ours $-$ & $\Delta$hIoU & $\Delta$hBIoU & $\Delta$p-hIoU \\
\midrule
Euler spiral & 0.042 {\scriptsize[0.030,0.055]} & 0.023 {\scriptsize[0.013,0.033]} & 0.040 {\scriptsize[0.028,0.052]} \\
Tangent & 0.048 {\scriptsize[0.036,0.061]} & 0.035 {\scriptsize[0.024,0.046]} & 0.044 {\scriptsize[0.032,0.057]} \\
LaMa + VT & 0.016 {\scriptsize[0.008,0.025]} & 0.030 {\scriptsize[0.021,0.041]} & 0.022 {\scriptsize[0.013,0.032]} \\
SDXL + VT & 0.230 {\scriptsize[0.201,0.260]} & 0.112 {\scriptsize[0.094,0.132]} & 0.215 {\scriptsize[0.187,0.244]} \\
FLUX Fill + VT & 0.109 {\scriptsize[0.088,0.130]} & 0.072 {\scriptsize[0.055,0.091]} & 0.110 {\scriptsize[0.091,0.129]} \\
Chord-closed & 0.107 {\scriptsize[0.087,0.128]} & 0.077 {\scriptsize[0.060,0.095]} & 0.095 {\scriptsize[0.075,0.116]} \\
VTracer & 0.175 {\scriptsize[0.148,0.203]} & 0.143 {\scriptsize[0.117,0.171]} & 0.233 {\scriptsize[0.202,0.266]} \\
LayerPeeler & 0.130 {\scriptsize[0.094,0.169]} & 0.076 {\scriptsize[0.048,0.105]} & 0.136 {\scriptsize[0.099,0.175]} \\
\bottomrule
\end{tabular}
\caption{Paired whole-image gains of \method{} (200 images, 732 targets)
with 95\% image-clustered bootstrap intervals, computed from unrounded
scores. VT: VTracer.}
\label{tab:paired-wholeimage}
\end{table}

\paragraph{Commonly matched targets.}
Table~\ref{tab:common-matched} averages over the 525 targets (188
images) matched by every shared-front-end method. On this subset the gain
over Euler is 0.059 $[0.043,0.075]$ hIoU, and over LaMa + VTracer 0.023
$[0.011,0.035]$ hIoU and 0.042 $[0.029,0.056]$ hBIoU.

\begin{table}[h]
\centering\small
\begin{tabular}{lrrrr}
\toprule
Method & hIoU & hBIoU & p-hIoU & Full IoU \\
\midrule
Chord closure & 0.427 & 0.161 & 0.406 & 0.722 \\
Tangent & 0.509 & 0.220 & 0.477 & 0.737 \\
Euler spiral & 0.518 & 0.237 & 0.483 & 0.728 \\
LaMa + VTracer & 0.554 & 0.226 & 0.507 & 0.736 \\
\textbf{Ours} & \textbf{0.577} & \textbf{0.268} & \textbf{0.539} & \textbf{0.760} \\
\bottomrule
\end{tabular}
\caption{Means over the 525 targets matched by every shared-front-end
method.}
\label{tab:common-matched}
\end{table}

\begin{figure}[h]
\centering
\includegraphics[width=\linewidth]{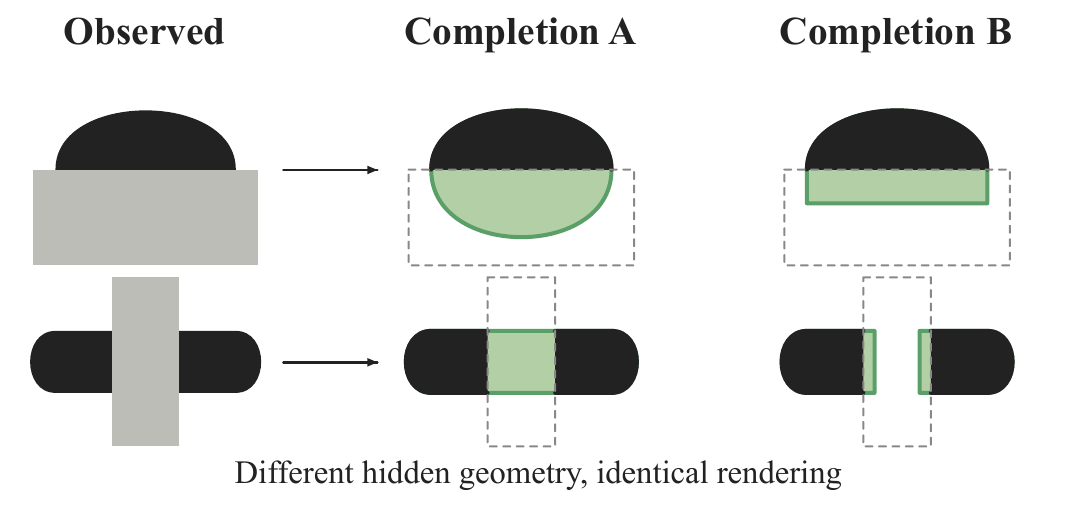}
\caption{Different hidden contours can agree with the same observed
composite. Schematic.}
\label{fig:ambiguity}
\end{figure}

\paragraph{Ambiguity of authored hidden geometry.}
Different hidden contours can produce the same composite
(Fig.~\ref{fig:ambiguity}), so we report both agreement with the
authored geometry and appearance after a layer is revealed.

\section{Controlled single-gap protocol}
\label{sec:controlled-protocol}
The controlled benchmark has 3,062 authored gap queries from 1,298
MMSVG-Illustration sources and 2,636 objects, drawn from three authored
pools of 877, 1,615, and 570 queries. Every query's authored hidden
interval deviates from its chord by more than 1.5 pixels; near-chord gaps
are scored separately in the KEEP analysis. Every method in
Table~\ref{tab:story-controlled} is scored on every query. Only the queried gap is
replaced; the object's other authored hidden intervals stay fixed as
common context, so the hidden-area and boundary metrics include them.

\paragraph{Common protection and fallback.}
The serialized SVG is checked against the supplied visible geometry and
occluder at $1024^2$ with a one-pixel tolerance. All methods share this
protection, the topology check against the chord base, and the chord
fallback. No reference contour is used to select a prediction. Final
chord fallback occurs on 5.6\% of queries for \method{}.

\paragraph{Pixel-completion adapters.}
LaMa~\cite{lama}, SDXL Inpainting~\cite{sdxl}, and FLUX.1
Fill~\cite{fluxfill} use the models, prompt, sampling
settings, and VTracer configuration of Sec.~\ref{sec:pixel-baselines},
with the object's visible shape in black on white as input and the query
occluder as the inpainting mask. An adapter then extracts the unique traced arc
between the supplied endpoints, using only the endpoints, visible
geometry, foreground coverage, and the prediction; if no unique arc
exists or assembly fails, the query falls back to the chord. Arc
extraction succeeds on 42.6\% (LaMa), 30.1\% (SDXL), and 38.5\% (FLUX
Fill) of queries, giving the final fallback rates of 59.0\%, 70.6\%, and
63.0\% in Table~\ref{tab:story-controlled}.

\begin{figure*}[t]
\centering
\includegraphics[width=\textwidth]{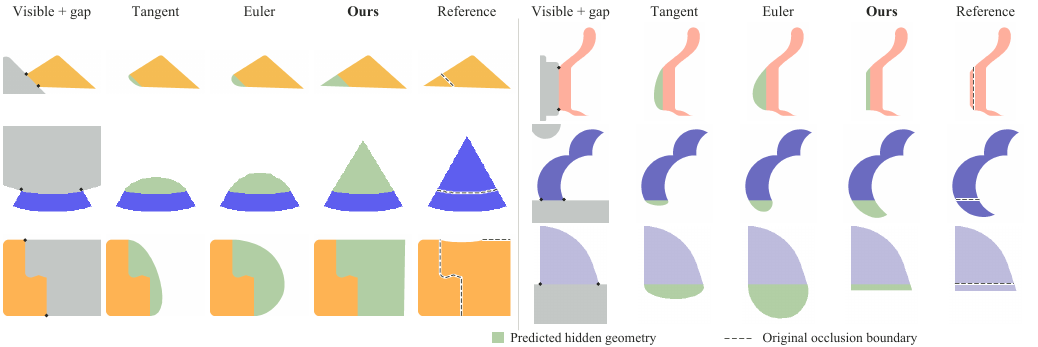}
\caption{\textbf{Controlled single-gap completions.} Selected queries with
corners or large turning angles; pale green marks predicted geometry inside the occluder and dots
mark the gap endpoints.}
\label{fig:controlled-corners}
\end{figure*}

\begin{figure*}[t]
\centering
\includegraphics[width=0.9\textwidth]{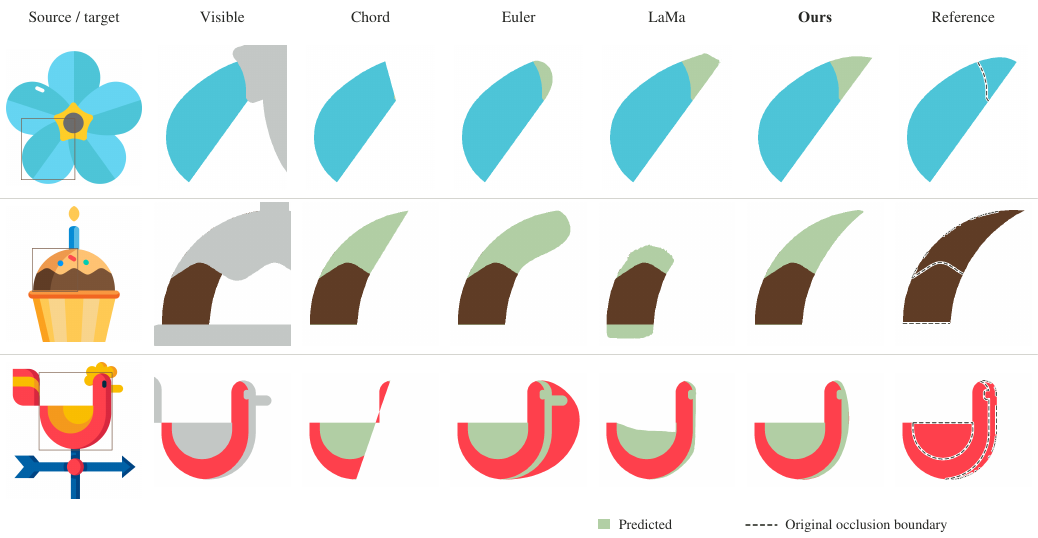}
\caption{\textbf{Completion from supplied visible curves and foreground geometry.} Source images provide visual context; they are not model inputs in this controlled experiment. Gray marks foreground coverage, pale green predicted hidden geometry, and dashed lines the original occlusion boundary on native-color references.}
\label{fig:qualitative-path}
\end{figure*}

\paragraph{Paired intervals for Table~\ref{tab:story-controlled}.}
Table~\ref{tab:paired-controlled} gives source-clustered paired
intervals (1,298 sources, 10,000 draws, seed 0). All exclude zero,
including those over Euler, tangent, and LaMa + VTracer. Stratified by
features of the reference gap arc, used only for grouping, the hIoU gain
over Euler is 0.034 $[0.021,0.045]$ on the 1,971 gaps with a corner
(turning above 30$^\circ$ between or within cubic segments) and 0.044
$[0.027,0.059]$ in the top tertile of maximum turning angle (2,000
source-clustered draws). Figure~\ref{fig:controlled-corners} shows such
queries, and Fig.~\ref{fig:qualitative-path} further completions from
supplied visible curves.

\begin{table}[h]
\centering\small
\setlength{\tabcolsep}{4pt}
\begin{tabular}{@{}lrr@{}}
\toprule
Ours $-$ & $\Delta$hIoU & $\Delta$hBIoU \\
\midrule
Euler spiral & 0.028 {\scriptsize[0.019,0.037]} & 0.038 {\scriptsize[0.028,0.048]} \\
Tangent & 0.060 {\scriptsize[0.051,0.069]} & 0.063 {\scriptsize[0.053,0.073]} \\
LaMa + VT & 0.124 {\scriptsize[0.112,0.137]} & 0.125 {\scriptsize[0.113,0.138]} \\
SDXL Inp.\ + VT & 0.202 {\scriptsize[0.191,0.212]} & 0.184 {\scriptsize[0.174,0.195]} \\
FLUX Fill + VT & 0.190 {\scriptsize[0.178,0.202]} & 0.177 {\scriptsize[0.165,0.189]} \\
Chord closure & 0.241 {\scriptsize[0.230,0.252]} & 0.227 {\scriptsize[0.216,0.238]} \\
\bottomrule
\end{tabular}
\caption{Paired controlled-gap gains of \method{} (3,062 queries) with
95\% source-clustered bootstrap intervals, computed from unrounded
scores.}
\label{tab:paired-controlled}
\end{table}

\paragraph{Timing.}
CPU times use a fixed subset of 100 queries, stratified over the three
pools and sampled with a fixed seed before any timing or inspection of
outputs. Every timed method processes the same queries on an Intel Xeon
Platinum 8360Y CPU with four threads, batch size one,
and one sequential process; method order rotates across queries, and
three warm-up queries per method are excluded. Table~\ref{tab:story-controlled}
reports the median over all 100 wall times, including fallbacks; the
means are 0.59, 1.08, 12.62, and 2.32\,s for tangent, Euler, the LaMa
pipeline, and \method{}. The timer covers prompt construction or input
rasterization, prediction, adaptation, common protection and assembly,
and SVG serialization (for LaMa, network inference, VTracer, and arc
extraction), and excludes model loading, disk I/O, and scoring.
Chord closure has no prediction step: its time consists of the shared
input protection, assembly, and serialization stages, whose median over
the same 100 queries is 0.60\,s in each of the four timed methods.
SDXL and FLUX Fill times are mean generation seconds per object, measured
in a single process on one unshared NVIDIA A100-SXM4 80GB GPU (2.45\,s
and 23.70\,s).

\section{Appearance after layer editing}
\label{sec:reveal-detail}
We use the 200-image collection and all 732 eligible targets. For each
target, the foreground layers above it are removed while the rest of the
scene is kept. Reveal metrics are computed in the reveal region, the union
of the reference foreground layers; CLIP-I compares a fixed target crop.
For an unmatched target, the unedited method rendering is used; a missing
output is a white canvas. Every target stays in the denominator.

\paragraph{Metric details.}
With RGB normalized to $[0,1]$, PSNR is
$-10\log_{10}(\max(\mathrm{MSE},10^{-10}))$ within the reveal region,
capped at 100\,dB and averaged over targets. SSIM uses an 11$\times$11
Gaussian window ($\sigma=1.5$), population covariance, and an RGB range of
255; we average the channel-mean SSIM map over the reveal region,
excluding a five-pixel image border, so windows can include surrounding
pixels. Reveal LPIPS uses the spatial VGG map. Crop CLIP-I is the cosine
similarity of normalized ViT-L/14 image embeddings of the reference
target box with 10\% padding, padded to a white square.

\paragraph{Paired differences.}
Resampling images 10,000 times (seed 0), \method{} reduces Reveal LPIPS
by 0.0103 $[0.0047,0.0165]$ relative to tangent continuation and by
0.1174 $[0.0983,0.1371]$ relative to VTracer.

\section{LayerPeeler configuration}
\label{sec:layerpeeler}
We use the released attention-based inference code and weights of
LayerPeeler~\cite{layerpeeler}. The
original Gemini 2.5 Pro interface was unavailable to us, so the
vision-language steps use \texttt{gemini-3-flash-preview} at thinking
level \texttt{low}, and VTracer replaces Recraft for vectorization (cutout
layers, filter speckle 8, length threshold 8, path precision 8; other
settings as in Sec.~\ref{sec:geometric-baselines}). Inputs are processed at
512 pixels, with seed 42, 40 diffusion steps, guidance 4.5, and at most
ten peeling steps.

To keep runs from aborting on malformed intermediate outputs, the
wrapper fills missing or non-PNG masks with their bounding box, reads four
coordinates under the \texttt{mask} key as a box, skips empty difference
layers, and re-requests malformed box-detection JSON up to three times
with an 8,192-token output limit.

All 200 icons and 732 targets are evaluated. Seven of the 200 runs
fail; failed runs and unmatched targets remain in the denominator with
zero recovery scores, white-canvas LPIPS, and zero size. The time in the
main whole-image table is the mean logged latency over the 178 images
with complete timing logs (median 98.5\,s). It covers the vision-language
calls, layer generation, retries within the run, frame extraction,
vectorization, and SVG assembly, and excludes model loading and input
resizing.

\section{Composite fidelity, representation size, and runtime}
\paragraph{Measurement.}
We render each final SVG at $1024^2$ on white and compute full-image
VGG LPIPS against the input raster, which measures the composite rather
than a visible-region mask. Size counts include hidden geometry and use
the same primitive-expansion convention for every method. An open
subpath has one more anchor than segments, and a closed subpath has equal
counts. Table~\ref{tab:representation-size} lists segments; the main
whole-image table lists anchors, which differ only for StarVector (220.6)
and OmniSVG (202.9). Both measures are image means over
the 200-image collection; failed outputs count zero.

The LaMa + VTracer composites replace each queried object of the
shared-front-end scene with its traced LaMa output; unqueried objects keep
the shared-front-end geometry. Its LPIPS and size describe these composites.
The mean SVG size of \method{} is 6.4\,KiB.

\begin{table}[t]
\centering\small
\setlength{\tabcolsep}{3pt}
\begin{tabular}{@{}lrr@{}}
\toprule
Method & \shortstack{Segments\\mean} & \shortstack{Segments\\median} \\
\midrule
VTracer & 476.2 & 430 \\
LayerPeeler & 339.3 & 262 \\
StarVector-8B & 217.9 & 75 \\
OmniSVG-1.1-8B & 202.4 & 193.5 \\
Chord-closed base & 100.4 & 92 \\
Tangent & 101.9 & 94 \\
Euler spiral & 105.1 & 96 \\
LaMa + VTracer & 969.9 & 535 \\
SDXL + VTracer & 664.6 & 604 \\
FLUX Fill + VTracer & 1604.1 & 1036 \\
\textbf{Ours} & 110.0 & 99 \\
\bottomrule
\end{tabular}
\caption{\textbf{Representation size} on the 200-image collection,
including visible and hidden geometry. Failed outputs count zero.
}
\label{tab:representation-size}
\end{table}

\paragraph{Runtime.}
All shared-front-end methods use the same front end, 94.44\,s per image
on average. Completion then adds 2.4\,s for \method{}, 31.8\,s for
LaMa + VTracer, 14.0\,s for SDXL + VTracer, and 134.0\,s for FLUX Fill +
VTracer (Sec.~\ref{sec:pixel-baselines}); chord, tangent, and Euler
continuation are not timed separately. VTracer alone takes 0.2\,s.
StarVector and OmniSVG times are means of generation and native export per
image, excluding model loading, with batch size one on A100 80GB GPUs;
StarVector ranges from 18.1 to
1,001.2\,s per image. LayerPeeler timing is described in
Sec.~\ref{sec:layerpeeler}.

\section{Component studies}
\label{sec:ablations-detail}
The whole-image component study uses the same 200 icons and 732 reference
targets as the main comparison, with shared assembly and the same
checkpoint; all variants retain every eligible target
(Table~\ref{tab:ablation-complete}).

\begin{table}[t]
\centering\small
\setlength{\tabcolsep}{2pt}
\begin{tabular}{@{}lrrrrr@{}}
\toprule
Variant & Matched & hIoU$\uparrow$ & hBIoU$\uparrow$ & p-hIoU$\uparrow$ & \shortstack{Full\\IoU$\uparrow$} \\
\midrule
Initial fit only & 527 & 0.211 & 0.068 & 0.200 & 0.485 \\
Chord closure & 526 & 0.306 & 0.115 & 0.291 & 0.518 \\
No sampling & 526 & 0.410 & 0.190 & 0.383 & 0.544 \\
No occluder input & 526 & 0.398 & 0.186 & 0.371 & 0.540 \\
KEEP disabled & 527 & 0.413 & 0.189 & 0.385 & 0.545 \\
\textbf{Ours} & 527 & \textbf{0.414} & \textbf{0.192} & \textbf{0.386} & \textbf{0.546} \\
\bottomrule
\end{tabular}
\caption{\textbf{Component variants} on 200 images and 732 targets. No sampling disables the sampled recovery proposals while keeping the deterministic fallbacks.}
\label{tab:ablation-complete}
\end{table}

\paragraph{Cross-frontend completion.}
We replace the anchor-guided paths with VTracer~\cite{vtracer} paths or AdaVec's~\cite{adavec} curve
stages on the same observed object masks. Object assignments, colors,
drawing order, foreground coverage, the frozen LM, and assembly are held
fixed. Each front end is evaluated with chord closure and with the same
LM plus sampling recovery. All variants cover 200 images and
732 reference targets (Table~\ref{tab:cross-frontend}).

Paths are assigned by visible-footprint IoU $\geq0.5$; portions within
1.5 pixels of foreground coverage are treated as hidden, and intervals
shorter than 3 pixels are merged. AdaVec uses its tracing, adaptive
simplification, and point-optimization stages on the shared masks, with
200 point-optimization steps. Native segmentation and joint whole-image
optimization are not used.

The AdaVec short-segment threshold is selected from $\{5,10,20,40\}$
on ten fixed input-order images using visible-fit quality and the input
anchor budget, without completion scores or hidden references. The selected
threshold is 40. The median per-object visible-chain anchor ratio is
1.50 relative to our front end, compared with 2.39 for VTracer.
With AdaVec's native threshold of 10, the median ratio is 2.00 and the
context contains 2,567 gap queries; Table~\ref{tab:cross-frontend}
reports this setting as well.
AdaVec-40 supplies more fragmented context: 2,987 gap queries versus
2,058 for VTracer and 1,772 for anchor-guided paths.

\begin{table*}[t]
\centering\small
\setlength{\tabcolsep}{4pt}
\begin{tabular}{llrrrrrr}
\toprule
Front end & Completion & Matched & hIoU$\uparrow$ & hBIoU$\uparrow$ & p-hIoU$\uparrow$ & Full IoU$\uparrow$ & LPIPS$\downarrow$ \\
\midrule
Anchor-guided fit & Chord & 526 & 0.306 & 0.115 & 0.291 & 0.518 & 0.0059 \\
Anchor-guided fit & LM & 527 & 0.414 & 0.192 & 0.386 & 0.546 & 0.0059 \\
VTracer paths & Chord & 516 & 0.282 & 0.085 & 0.265 & 0.503 & 0.0100 \\
VTracer paths & LM & 516 & 0.359 & 0.138 & 0.330 & 0.520 & 0.0099 \\
AdaVec, threshold 40 & Chord & 513 & 0.261 & 0.065 & 0.241 & 0.492 & 0.0216 \\
AdaVec, threshold 40 & LM & 513 & 0.316 & 0.100 & 0.289 & 0.504 & 0.0215 \\
\midrule
AdaVec, threshold 10 (native) & Chord & 513 & 0.252 & 0.063 & 0.234 & 0.490 & 0.0166 \\
AdaVec, threshold 10 (native) & LM & 513 & 0.316 & 0.107 & 0.289 & 0.505 & 0.0165 \\
\bottomrule
\end{tabular}
\caption{\textbf{Paired completion across three curve front ends.}
All geometry means include the same 732 targets; composite LPIPS averages
200 images. LM uses the same frozen weights, two cyclic passes, and
sampling recovery for every front end. The last two rows use AdaVec's
native short-segment threshold instead of the selected value.}
\label{tab:cross-frontend}
\end{table*}

Paired LM-minus-chord hIoU gains are 0.077 $[0.060,0.095]$ on VTracer
paths and 0.056 $[0.042,0.070]$ on AdaVec paths; the corresponding
p-hIoU gains are 0.066 $[0.050,0.084]$ and 0.048 $[0.034,0.062]$. The
gain on our own paths is given in Table~\ref{tab:paired-wholeimage}.
With the native AdaVec threshold, the gains are 0.064 $[0.049,0.081]$
in hIoU and 0.056 $[0.042,0.072]$ in p-hIoU. Final hIoU rounds to
0.316 for both threshold settings.
Intervals use 10,000 image-clustered bootstrap draws with seed 0.

\paragraph{Completion choices.}
Initial fit only keeps the fitted object without any gap closure.
Chord closure closes every gap with its chord under the same assembly.
No-sampling recovery retains tangent, chord, and initial-fit fallbacks
while disabling the eight sampled proposals. These variants keep the
checkpoint fixed.

Relative to the initial fit, the full system gains 0.202 $[0.173,0.232]$
in hIoU and 0.125 $[0.103,0.148]$ in hBIoU. Sampling recovery adds
0.0030 $[0.0006,0.0065]$ hIoU and 0.0021 $[0.0005,0.0043]$ hBIoU.
Removing it changes the output of 27 of the 200 images.

\paragraph{Occluder input and KEEP.}
These fixed-checkpoint interventions retain the same predicted front end,
final-SVG matching rule, assembly masks, and deterministic fallbacks.
No occluder input removes the entire foreground-token block from single-gap,
paired, iterative, and recovery-sampling prompts.
KEEP disabled forbids KEEP in every greedy query. Recovery sampling already
excludes KEEP and retains its seeds, eight proposals, and temperature of 0.7;
deterministic fallback can still produce a chord.

Full-system minus no-occluder gains are 0.0155 $[0.0086,0.0227]$ hIoU,
0.0066 $[0.0004,0.0135]$ hBIoU, 0.0152 $[0.0081,0.0226]$ p-hIoU,
and 0.0059 $[0.0018,0.0103]$ full IoU.
The full-system minus KEEP-disabled gain is 0.0031 $[0.0007,0.0059]$
hBIoU.
All intervals use 10,000 paired image-clustered bootstrap draws, seed 0.

\paragraph{Controlled gaps that need no extension.}
The 3,062 queries of the controlled table exclude, by construction, gaps
whose authored interval lies within 1.5 pixels of the chord. The authored
controlled pools contain 672 such gaps from 342 sources
(Table~\ref{tab:keep-subset}). They are scored separately from the controlled table. Excess is the predicted area inside
the occluder but outside the reference hidden region, divided by the
reference hidden area. The full model outputs KEEP on 86.8\% of these
gaps.

\begin{table}[h]
\centering\small
\setlength{\tabcolsep}{3pt}
\begin{tabular}{lrrrr}
\toprule
Method & hIoU & Excess & Excess$>$10\% & hIoU$<$0.9 \\
\midrule
Chord closure & 0.999 & 0.03\% & 0.0\% & 0.0\% \\
Tangent & 0.971 & 6.0\% & 7.4\% & 7.4\% \\
Euler spiral & 0.966 & 11.4\% & 7.4\% & 8.2\% \\
KEEP disabled & 0.904 & 10.7\% & 15.0\% & 27.1\% \\
\textbf{Ours} & 0.972 & 5.8\% & 6.5\% & 7.1\% \\
\bottomrule
\end{tabular}
\caption{\textbf{The 672 authored controlled gaps within 1.5 pixels of
the chord.} Excess: mean excess-area fraction.}
\label{tab:keep-subset}
\end{table}

Source-clustered paired differences (10,000 draws, seed 0) are, for the
full model minus KEEP disabled, 0.067 $[0.057,0.079]$ hIoU and $-0.050$
$[-0.082,-0.027]$ excess fraction. Chord closure exceeds tangent by 0.028
$[0.016,0.041]$ and Euler by 0.033 $[0.020,0.050]$ hIoU.

\section{User-study protocol}
\label{sec:user-study-protocol}
\begin{figure*}[t]
\centering
\includegraphics[width=0.9\textwidth]{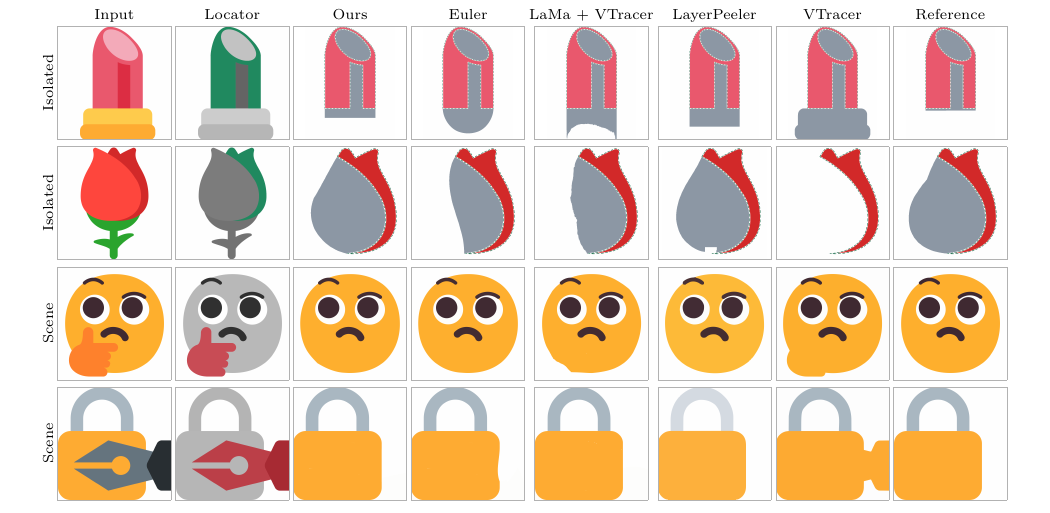}
\caption{\textbf{Example user-study items}, two isolated targets (top)
and two edited scenes (bottom), with method names revealed; participants
saw the candidates under letters. Isolated targets: color is the
originally visible part, gray everything else, dashed the visible
contour. Scenes: the foreground marked in the locator is removed.}
\label{fig:user-study-examples}
\end{figure*}

The blinded rating study uses 14 distinct source images: eight
isolated-target items and six edited scenes after foreground removal
(Fig.~\ref{fig:user-study-examples}). Each item shows the input, a
locator, and anonymized candidates under letters, including \method{},
LaMa + VTracer, Euler continuation, LayerPeeler, VTracer without
completion, and the authored reference. Candidate order is randomized across
items. Participants
rate how complete, coherent, and natural each displayed shape appears on
a five-point scale (1: severe problems; 5: complete and natural); ties
and an ``unable to judge'' answer are allowed. The two viewing conditions
are analyzed separately.

In isolated-target items, the locator marks the visible part of the
target; every candidate shows the originally visible part in its input
color and everything else in a common gray, including spill, with the
visible contour dashed. Scene items mark the foreground to be moved and
show each candidate's remaining scene in its own colors after removing
the corresponding native elements. Locators expose only visible support derived from the
reference layering, never hidden contours.

\paragraph{Responses and aggregation.}
Eighteen participants each completed one sheet, with no missing answers.
For each candidate and condition we average 144 ratings
over the eight isolated-target items and 108 over the six scenes.

\paragraph{Mean ratings.}
The mean ratings for isolated targets / edited scenes are 2.00/2.23 for
VTracer, 2.71/3.26 for LaMa + VTracer, 2.99/2.96 for LayerPeeler,
3.38/3.82 for Euler continuation, 3.84/4.34 for \method{}, and 3.58/4.91
for the authored reference.

\section{Baseline configurations and comparisons}
\label{sec:planned}
Whole-image methods share the 200-icon evaluation collection; the
hosted-model curve comparison uses its own 20 fixed queries.

\subsection{Tracing and geometric continuation}
\label{sec:geometric-baselines}
\paragraph{VTracer.}
The tracing baseline runs VTracer 0.6.11 on the $1024^2$ input in color
mode with stacked layers and spline curves (filter speckle 4, color
precision 6, layer difference 16, corner threshold 60, length threshold 4,
ten iterations, splice threshold 45, path precision 3).

\paragraph{Chord, tangent, and Euler-spiral continuation.}
Chord closure joins the two gap endpoints with a straight segment. Tangent
continuation uses one cubic B\'ezier whose end tangents continue the
adjacent visible segments, with both handles set to one third of the chord
length; a degenerate tangent falls back to the chord direction. The
Euler-spiral baseline~\cite{eulerspiral} fits a single spiral segment,
whose curvature varies linearly with arc length, through both endpoints
and their tangents by one-dimensional root finding, keeping the solution
with the smallest total turning. The spiral is sampled at 200 points and
converted to one to three cubic B\'ezier segments. If no root exists, the total turning exceeds $2\pi$, or
the arc is longer than five chord lengths, the tangent continuation is
used.

\subsection{Pixel completion with LaMa, SDXL Inpainting, and FLUX.1 Fill}
\label{sec:pixel-baselines}
\paragraph{Shared pixel-to-vector adapter.}
The three pixel baselines differ only in the inpainting model. Each
queried object is given as a $1024^2$ RGB image of its predicted visible
shape, black on white, together with its predicted foreground coverage as
the inpainting mask. Known pixels are kept, $\mathrm{mask}\cdot
\mathrm{pred}+(1-\mathrm{mask})\cdot\mathrm{input}$, and the result is
binarized at mean RGB $<0.5$. VTracer 0.6.11 (binary mode, spline
curves; filter speckle 4, corner threshold 60, length threshold 4, ten
iterations, splice threshold 45, path precision 3) traces the completed
shape. In the whole-image comparison, the traced object replaces the
queried object in the shared-front-end scene; unqueried objects keep the
initial fit, and no chord fallback is applied. In the controlled
comparison, the same traced output passes through the gap adapter, common
protection, and chord fallback of Sec.~\ref{sec:controlled-protocol}.
LaMa uses the big-LaMa model in FP32 on an Intel Xeon Platinum 8360Y
CPU.

\paragraph{Diffusion inpainters.}
Both diffusion models use one fixed prompt, ``flat vector icon, solid
colors, black shape on white background'', no negative prompt, $1024^2$
resolution, one sample per object, and no candidate selection; the seed
is reset to 0 for every object.
SDXL Inpainting uses the public
\texttt{stable-diffusion-xl-1.0-\allowbreak inpainting-0.1} checkpoint
with \texttt{AutoPipeline\allowbreak ForInpainting} in FP16, 20 steps, strength 0.99,
and guidance 8.0.
FLUX.1 Fill uses \texttt{FLUX.1-Fill-dev} with
\texttt{FluxFillPipeline} in BF16, following the model-card example:
50 steps, guidance 30, maximum sequence length 512, and the pipeline's
default strength 1.0; the pipeline does not accept a negative prompt.
All generations succeed in both comparisons.

\paragraph{Timing.}
The SDXL and FLUX.1 Fill completion times in the whole-image table sum
per-object generation and VTracer time for each image and average over
the 178 images with complete timing logs. Generation is measured with
one process on an otherwise idle NVIDIA A100-SXM4 80GB: 2.45\,s per
object for SDXL and 23.70\,s for FLUX.1 Fill. The controlled table reports these per-object GPU times.

\begin{figure*}[t]
\centering
\includegraphics[width=0.85\textwidth]{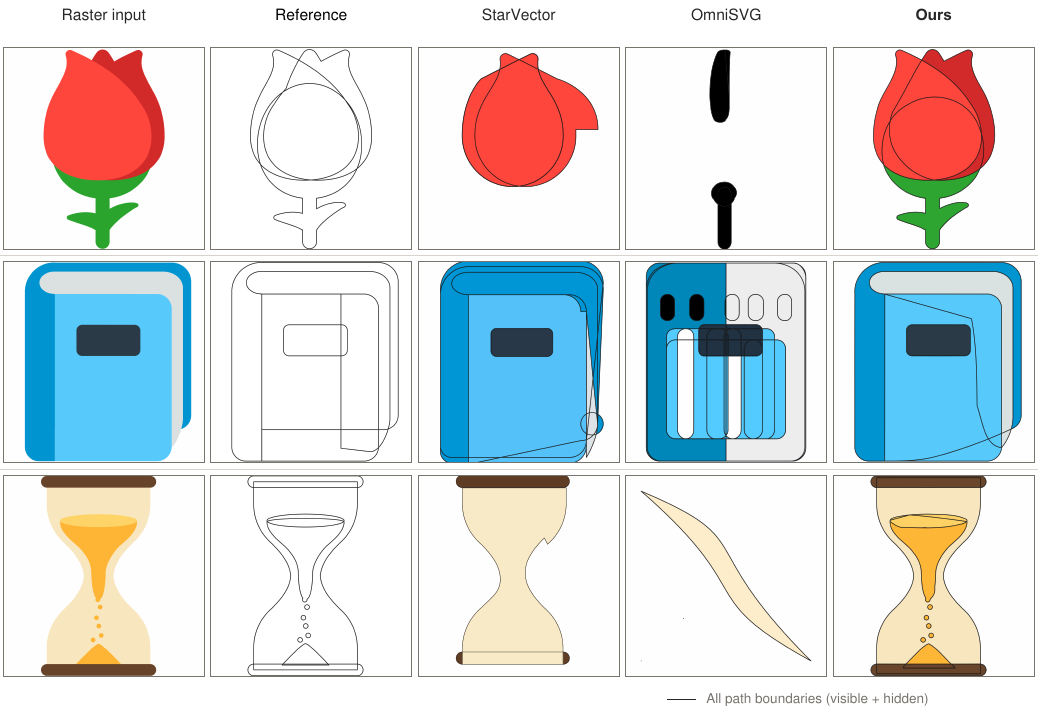}
\caption{\textbf{Native SVG generation and hidden-shape recovery.}
Raster inputs and authored references precede StarVector,
OmniSVG and \method{} outputs. Each output retains its own colors; solid
lines expose all path boundaries. The native image-to-SVG interfaces do
not receive an explicit completion instruction.}
\label{fig:native-svg}
\end{figure*}

\subsection{Independent SVG generation}
\paragraph{Models and inputs.}
StarVector-8B~\cite{starvector} uses
\texttt{starvector/starvector-8b-im2svg}; OmniSVG~\cite{omnisvg} uses
\texttt{OmniSVG/OmniSVG1.1\_8B}. Their released image-to-SVG interfaces
receive the original raster, resized only as required by the processor,
without an explicit amodal-completion instruction. Outputs retain their
predicted paint order and are not re-vectorized
(Fig.~\ref{fig:native-svg}).
StarVector uses maximum length 10,000, temperature 1.0, top-$p$ 0.95,
three beams, and length penalty 0.5, retaining one final candidate
through its native postprocessor. OmniSVG uses at most 1,536 new tokens,
temperature 0.3, top-$p$ 0.9, top-$k$ 50, repetition penalty 1.05, and
one beam. Both run in BF16 with a fixed per-input seed.

\paragraph{Scoring.}
Every eligible target remains in the denominator, including unmatched
targets and failed outputs. OmniSVG produces an output for all 200 inputs,
but nine cannot be parsed or rendered by the evaluator; they receive zero
recovery scores, white-canvas LPIPS, and zero size. On the 191 remaining
images alone, OmniSVG has LPIPS 0.2921 versus 0.0059 for \method{}, so
the appearance difference does not stem from these failures. All
StarVector outputs are evaluable.

\paragraph{Unavailable baselines.}
At the time of our experiments, SemLayer~\cite{semlayer} had released no inference code or
weights, and AmodalSVG~\cite{amodalsvg} stated that code would be released upon acceptance.

\subsection{Curve completion by hosted chat models}
\label{sec:chat-diagnostic}
\paragraph{Queries and inputs.}
The 20 source-distinct queries are drawn from the authored controlled-gap
pools by deterministic hashing and source deduplication before any model
is scored. Each model receives the quantized visible chain in the gap frame,
endpoints $A=(0,0)$ and $B=(32,0)$, a 16-sample foreground seam, and an
eight-sample depth profile; hidden reference geometry and semantic labels
are omitted. The instruction asks for the missing boundary from $A$ to
$B$, either as \texttt{KEEP} or as one path string of \texttt{L}/\texttt{C}
commands, without rendering, solver use, or revision. Queries are sent in
batches of five at medium reasoning effort,
and the first answer is scored. Our 107M curve model receives the same
geometric quantities as tokens; it is scored without the whole-system
sampling recovery.

\paragraph{Scoring.}
A deterministic parser checks syntax and endpoints without repairing
the geometry; valid intervals undergo common assembly and $1024^2$ raster
evaluation, and \texttt{KEEP} is scored as a straight closure. Five empty
responses (one Gemini batch) score zero. Table~\ref{tab:chat-full} lists
the scores.

\begin{table}[t]
\centering\small
\setlength{\tabcolsep}{4pt}
\begin{tabular}{@{}lrrr@{}}
\toprule
Model & Valid (\%) & p-hIoU$\uparrow$ & hBIoU$\uparrow$ \\
\midrule
GPT-6 Astra & 100 & 0.390 & 0.302 \\
GPT-6.1 Sol & 100 & 0.387 & 0.256 \\
Claude Fable 5.2 & 100 & 0.350 & 0.253 \\
Claude Opus 5.5 & 100 & 0.244 & 0.143 \\
Gemini 4 Argon & 75 & 0.308 & 0.212 \\
\textbf{Ours} (107M) & 100 & \textbf{0.421} & \textbf{0.322} \\
\bottomrule
\end{tabular}
\caption{\textbf{Hosted-model curve completion} on the 20 queries.
Failed answers score zero.}
\label{tab:chat-full}\label{tab:chat-paired}
\end{table}

\paragraph{Time and cost.}
Complete usage logs exist for the same three batches (15 queries) for all
five models. For each batch we sum the recorded request durations and
charges, divide by five, and average over the batches. We time our model
on the same 15 inputs with batch size one after warm-up, synchronizing the
GPU around each prediction: 0.30\,s on one NVIDIA A100 80GB. Its cost is
$t\cdot r/3600$ with the on-demand rental rate $r=\$2.09$ per GPU-hour
for a cloud server with that card, giving 0.018 cents per
query, excluding loading and idle time.

\subsection{Qualitative comparisons}
\paragraph{GPT-6 Astra in the qualitative figures.}
For the scenes shown in Figs.~\ref{fig:qualitative}
and~\ref{fig:approved-examples}, GPT-6 Astra
receives the full raster and a fixed instruction to generate a complete
layered SVG that retains all foreground objects. Each image uses an
independent session and the first answer, without repair or selection.
Isolated-target views extract the corresponding object from this
full-scene SVG; the model receives no target crop, visible curve, hidden
mask, or reference. These generations are qualitative only. OmniSVG is
omitted from the isolated-target figure because its outputs contain no
identifiable separate target in two of the four scenes. All displayed
outputs are saved original outputs without geometric repair.

\end{document}